\documentclass[10pt,twocolumn,letterpaper]{article}

\usepackage[pagenumbers]{cvpr} 

\usepackage{amsmath,amsfonts,bm}

\def\eqref#1{equation~\ref{#1}}

\def\1{\bm{1}}

\def\rmI{{\mathbf{I}}}

\def\rmK{{\mathbf{K}}}

\def\vc{{\bm{c}}}

\def\vn{{\bm{n}}}

\def\vz{{\bm{z}}}

\DeclareMathAlphabet{\mathsfit}{\encodingdefault}{\sfdefault}{m}{sl}
\SetMathAlphabet{\mathsfit}{bold}{\encodingdefault}{\sfdefault}{bx}{n}

\def\gD{{\mathcal{D}}}
\def\gE{{\mathcal{E}}}

\def\gL{{\mathcal{L}}}
\def\gM{{\mathcal{M}}}

\def\gU{{\mathcal{U}}}

\def\sR{{\mathbb{R}}}

\newcommand{\vx}{\boldsymbol{x}}
\newcommand{\vy}{\boldsymbol{y}}
\newcommand{\x}{\mathbf{x}}

\newcommand{\y}{\mathbf{y}}
\newcommand{\Y}{\mathbf{Y}}
\newcommand{\Wsij}{W^s_{ij}}
\newcommand{\Wc}{\mathbf{W}(\vc)}
\newcommand{\Ws}{\mathbf{W^s}}
\newcommand{\J}{\mathbf{J}}

\usepackage{enumitem}
\usepackage{xspace}
\usepackage{booktabs}
\usepackage{colortbl}
\usepackage{pifont}
\usepackage{graphicx}
\usepackage{siunitx}

\usepackage{caption}
\usepackage{algorithm}
\usepackage{algpseudocode}

\def\modelname{LatentCRF\xspace}

\definecolor{Gray}{gray}{0.90}

\newcommand{\cmark}{\ding{51}}
\newcommand{\xmark}{\ding{55}}

\newcommand{\inc}[1]{\ensuremath{_{\text{\textcolor{PineGreen}{(+#1)}}}}}
\newcommand{\dec}[1]{\ensuremath{_{\text{\textcolor{RedOrange}{(-#1)}}}}}

\usepackage{url}

\definecolor{cvprblue}{rgb}{0.21,0.49,0.74}
\usepackage[pagebackref,breaklinks,colorlinks,allcolors=cvprblue]{hyperref}

\def\paperID{5007} 
\def\confName{CVPR}
\def\confYear{2025}

\title{\modelname: Continuous CRF for Efficient Latent Diffusion}

\author{
Kanchana Ranasinghe$^{1,2}$, 
Sadeep Jayasumana$^{1}$, 
Andreas Veit$^{1}$, 
Ayan Chakrabarti$^{1}$, 
Daniel Glasner$^{1}$ 
\\
Michael S Ryoo$^{2}$ , 
Srikumar Ramalingam$^{1}$, 
Sanjiv Kumar$^{1}$ 
\vspace{0.5em} \\ 
$^{1}$Google Research \quad $^{2}$Stony Brook University \vspace{0.5em} \\
\texttt{kranasinghe@cs.stonybrook.edu} \\
}

\begin{document}


\twocolumn[{%
\renewcommand\twocolumn[1][]{#1}%
\maketitle
\begin{center}
    \setcounter{figure}{0}
    \centering
    \captionsetup{type=figure}
    \vspace{-1.5em}
    \includegraphics[width=1.0\textwidth,trim={0.0cm 0.0cm 0cm 0},clip,page=1]{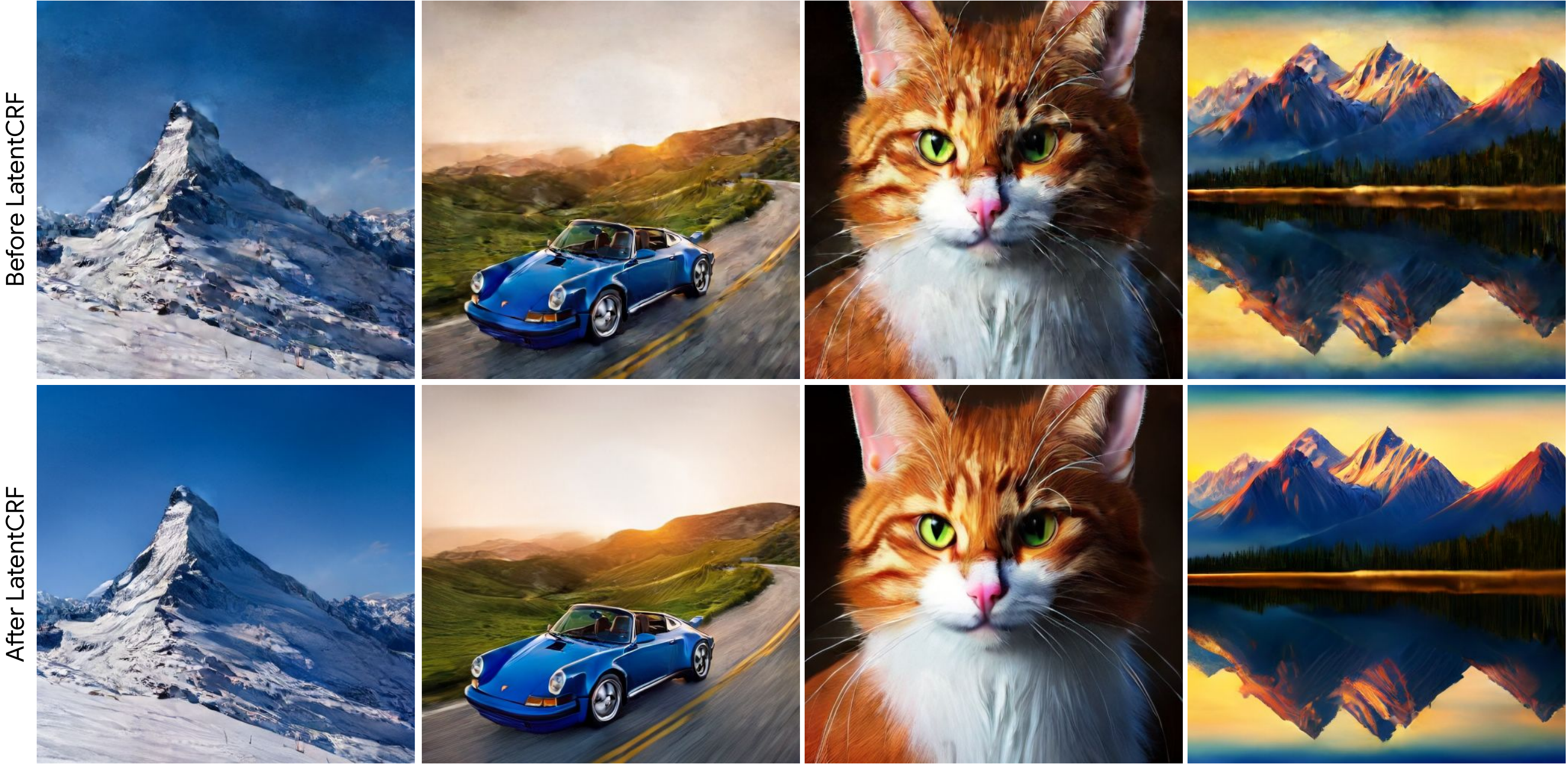}\vspace{-0.5em}%
    \caption{\textbf{Speeding up LDM inference with LatentCRF:} The figure shows the effect of applying our proposed lightweight Conditional Random Field (CRF) in the latent space of a Latent Diffusion Model (LDM). By replacing several LDM iterations with a single LatentCRF inference, we are able to increase inference speed by 33\%, while maintaining image quality and generation diversity.}\label{fig:teaser}%
    \vspace{1.0em}
\end{center}%

}]

\maketitle

\begin{abstract}
Latent Diffusion Models (LDMs) produce high-quality, photo-realistic images, however, the latency incurred by multiple costly inference iterations can restrict their applicability. We introduce \modelname, a continuous Conditional Random Field (CRF) model, implemented as a neural network layer, that models the spatial and semantic relationships among the latent vectors in the LDM. By replacing some of the computationally-intensive LDM inference iterations with our lightweight \modelname, we achieve a superior balance between quality, speed and diversity. We increase inference efficiency by 33\% with no loss in image quality or diversity compared to the full LDM. \modelname is an easy add-on, which does not require modifying the LDM.
\vspace{-0.5em}
\end{abstract}

\section{Introduction}

Recent text-to-image methods are remarkably successful in generating photo-realistic images that are faithful to the input text prompts~\cite{saharia2022photorealistic,Rombach2021HighResolutionIS,yu2022scaling,ramesh2022hierarchical,Midjourney2022}. They outperform Generative Adversarial Networks (GANs)~\cite{goodfellow2014generative} in terms of quality, the complexity of text prompts they can handle, and training stability (avoiding mode collapse). However, these breakthroughs come with a cost. Unlike single-step GANs, most existing algorithms require multiple iterations and thus incur significant inference cost.

\begin{figure*}[t]
    \begin{minipage}{\linewidth}
        \centering\centering
        \includegraphics[width=\linewidth]{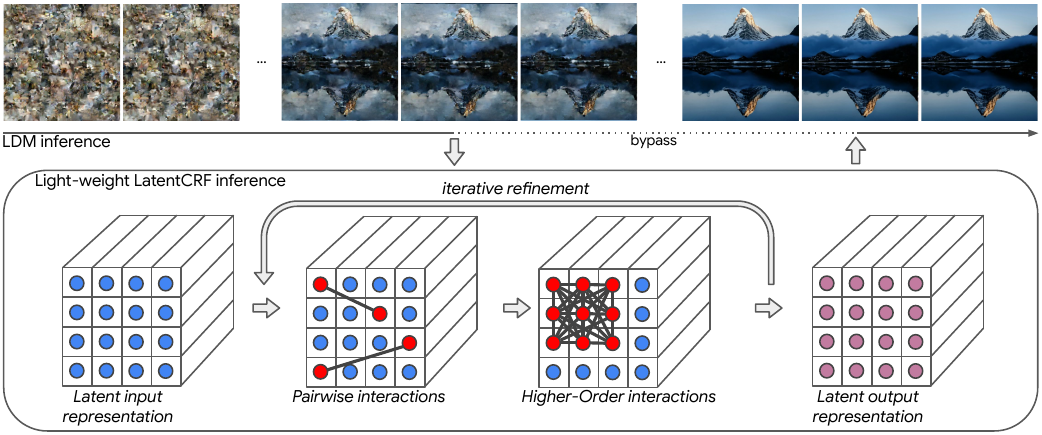}
    \end{minipage}
    \vspace{-0.5em}
    \caption{
    \textbf{Overview of LatentCRF:} 
    We replace several LDM inference iterations with an application of our CRF in the LDM's latent space. 
    Our \modelname modifies latent vectors with pairwise and higher-order interactions to better align with the distribution of natural image latents. The CRF inference cost is insignificant compared to the LDM's U-Net, leading to significant savings in inference time.
    }
    \label{fig:method}
\end{figure*}

Diffusion models have become the default choice for image generation.
An important idea in Latent Diffusion Models (LDMs) \cite{Rombach2021HighResolutionIS} is that significant savings can be obtained by denoising in a lower-dimensional latent space, instead of the original pixel space. Other ways of improving the speed of LDMs include distillation to reduce the number of inference steps~\citep{salimans2022progressive}, and model compression methods~\citep{choi2023squeezing, li2023snapfusion}. 

In this work, we take an alternative approach. We speed up diffusion using a continuous Conditional Random Field (CRF) model formulated as a trainable neural network layer~\cite{Zheng2015ConditionalRF,Jayasumana2023MarkovGenSP}. Our continuous CRF layer is an order of magnitude less expensive than the LDM U-Net. Our method is easy to apply, in that it can be trained with relatively few resources, and doesn't require modifying the LDM. Moreover, we could potentially combine it with other efficiency improving methods like model compression.

\begin{figure*}[t]
    \centering
    \begin{minipage}{0.03\linewidth}
    \centering
    \begin{tabular}{c}
         \rotatebox{90}{\texttt{LDM}}   \\[3.8em] 
         \rotatebox{90}{\texttt{ADD}}   \\[3.8em] 
         \rotatebox{90}{\texttt{Ours}}  \\
    \end{tabular}
    \end{minipage}
    \begin{minipage}{0.96\linewidth}
    \begin{minipage}{\linewidth}
        \centering\centering
        \includegraphics[width=\linewidth]{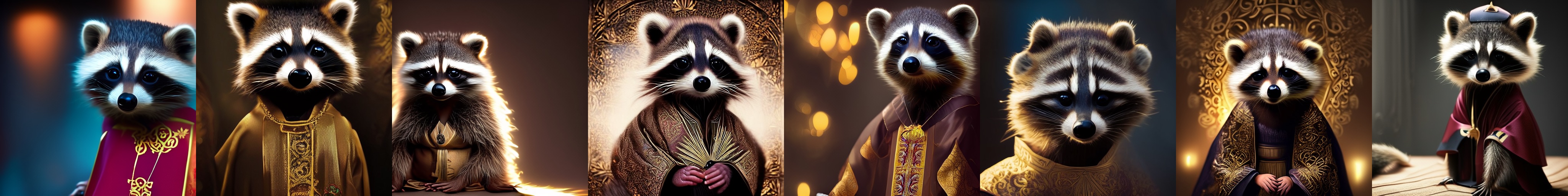}
    \end{minipage}
    \begin{minipage}{\linewidth}
        \centering
        \includegraphics[width=\linewidth]{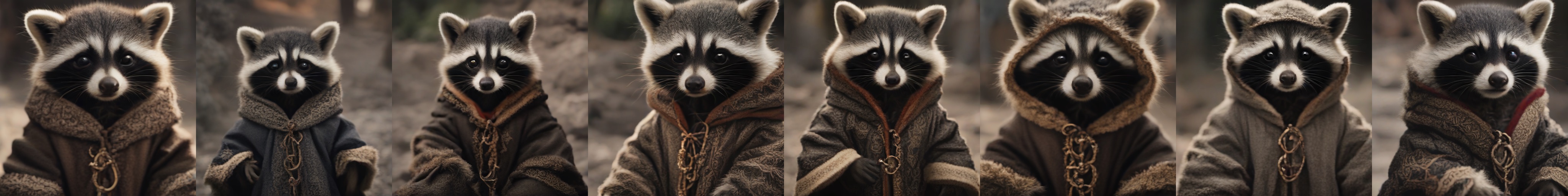}
    \end{minipage}
    \begin{minipage}{\linewidth}
        \centering\centering
        \includegraphics[width=\linewidth]{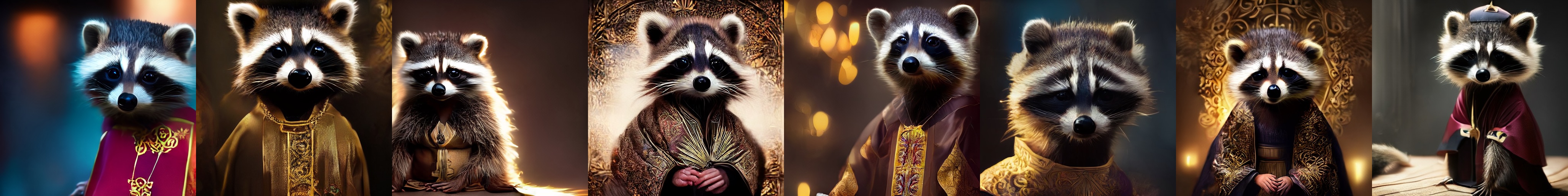}
    \end{minipage}
    \end{minipage}
    \vspace{-0.5em}
    \caption{
    \textbf{Diversity of generations:} We generate multiple images for the prompt \texttt{`A cinematic shot of a baby racoon wearing an intricate italian priest robe.'} with varying input noise. We observe that our LatentCRF (bottom) retains the diversity of LDM~\cite{Rombach2021HighResolutionIS} teacher (top). In contrast, distillation-based approaches like SDXL-Turbo~\citep{Sauer2023AdversarialDD} (middle) tend to suffer from decreased diversity (more details in \Cref{table:diversity}). 
    }
    \label{fig:diversity}
\end{figure*}

When evaluating efficient diffusion models it's important to pay attention to \emph{quality}, \emph{speed} and \emph{diversity}. As noted in \cite{Sauer2023AdversarialDD} and demonstrated in our experiments (\Cref{table:diversity}), distillation methods that reduce the number of steps are susceptible to \emph{diversity losses}, given a text prompt and different noise image inputs they tend to produce very similar outputs (\Cref{fig:diversity}). On the other hand, distillation methods which compress the network are susceptible to quality losses (\Cref{table:res_compress}). We believe that our method strikes a good balance. As we show in \Cref{sec:experiments} it increases inference speed by 33\%, with no quality loss (\Cref{table:results_main}, \Cref{fig:qualitative_results}) and virtually no diversity loss (\Cref{table:diversity}, \Cref{fig:diversity}). 

The most relevant prior work is that of~\cite{Jayasumana2023MarkovGenSP} which speeds up text-to-image with an MRF in a discrete token space. In contrast, our approach i) works with continuous latent features rather than discrete, quantized token labels, ii) introduces higher-order energy terms derived using a Field of Experts model~\cite{Roth2005FieldsOE}, and iii) uses a CRF model instead of MRF with text conditioning in the pairwise energy terms.

In our LatentCRF training, we learn the distribution of natural images with a denoising and adversarial loss, followed by an LDM schedule-aware distillation training that integrates the continuous CRF with the LDM inference pipeline. 
In summary, our key contributions are,
\begin{enumerate}
[leftmargin=3.0em,noitemsep,topsep=-0.4em,itemsep=-0.7ex,partopsep=0ex,parsep=1ex]
    \item We propose a novel \modelname model with unary, pairwise, and higher-order energy terms to learn the spatial and semantic relationships among latent vectors representing natural images. 
    \item We design a two-stage training algorithm with novel objectives for learning the \modelname model as a neural network layer.
    \item By replacing some LDM iterations with \modelname, we improve inference speed by 33\% with no quality or diversity loss.
    \item \modelname is an easy add-on to existing LDMs. It can be trained quickly, and with fewer resources, and doesn't require modifying the LDM.
\end{enumerate} 
\vspace{0.5em}

\noindent

\section{Related Work}

\noindent{\bf Text-to-Image Generation:} 
We are seeing a renaissance in the field of text-to-image generation~\citep{chang2023Muse,saharia2022photorealistic,ramesh2022hierarchical,Rombach2021HighResolutionIS,yu2022scaling,zhang2023adding,singer2022makeavideo,villegas2022phenaki,ge2022long,kumari2023multiconcept}. This breakthrough has been possible through multiple innovations: (1) joint image-text embedding models~\cite{radford2021learning}, (2) image encoders and decoders~\cite{Oord2017,esser2021taming}, (3) the emergence of diffusion models~\cite{pmlr-v37-sohl-dickstein15}, and (4) token-based methods~\cite{yu2022scaling}.

\noindent{\bf Diffusion models:} 
The idea of diffusion models is to gradually convert one distribution to another, typically a complex unknown data distribution to a simple known Gaussian one. The forward process systematically and slowly destroys structure in data by adding noise, and the reverse process learns to restore the original data through multiple denoising steps~\cite{pmlr-v37-sohl-dickstein15}. This simple, yet powerful idea, has been one of the breakthrough techniques used in image generation~\cite{saharia2022photorealistic,ramesh2022hierarchical,Rombach2021HighResolutionIS,Nichol2022,Midjourney2022,gafni2022makeascene}.
Diffusion does not suffer from many of the limitations of other methods, e.g., quantization artifacts in discrete token-based methods~\cite{yu2022scaling}. However, even multi-resolution methods, where a coarse image is generated by a faster module and superresolved with a more expensive module~\cite{saharia2022photorealistic}, still require numerous denoising steps making inference costly.

\noindent{\bf Latent Diffusion Models:}
By performing the diffusion process in a lower-dimensional latent space instead of the original pixel space, latent diffusion models~\cite{Rombach2021HighResolutionIS} achieve significant improvement in training and inference efficiency.
Combined with efficient sampling strategies \citep{Song2020DenoisingDI}, LDMs exhibit notable inference speed-ups over pixel-space Denoising Diffusion Probabilistic Model (DDPM)~\citep{Ho2020} while mostly retaining both visual quality and diversity of generated images \citep{Rombach2021HighResolutionIS}.

\noindent
\textbf{Beyond LDMs:}
Techniques speeding up LDMs can be broadly classified into two categories. First, there are many techniques that focus on efficient sampling, i.e. searching for a more efficient path that goes from noise to data with fewer steps~\citep{liu2022flowstraightfastlearning,Esser2024ScalingRF,Song2020DenoisingDI,liu2022pndm,jolicoeurmartineau2021gotta,karras2022elucidating,dockhorn2022genie,zhang2022gddim,zhao2023unipc,lu2022dpm++,lu2022dpm,pernias2023wuerstchenefficientarchitecturelargescale}.

Second, are distillation approaches. Distillation can be ``temporal", where the teacher and student use the same architecture and the student is trained to generate images with fewer steps~\citep{salimans2022progressive}~\citep{Sauer2023AdversarialDD}, or ``compression" where a student with fewer parameters is trained to mimic a teacher with more. For the latter, the student architecture is usually restricted to variations of the popular U-Net architecture, to allow weight initialization from a pre-trained diffusion model \citep{choi2023squeezing, li2023snapfusion,Habibian2023ClockworkDE,kim2023architectural,dockhorn2023distilling,Ma2023DeepCacheAD,Wimbauer2023CacheMI,Zhao2023MobileDiffusionIT,Tang2024AdaDiffAD}. 
While distillation obtains good inference speedups, methods tend to trade-off either diversity or photo-realism~\citep{Sauer2023AdversarialDD,luhman2021knowledge,salimans2022progressive,liu2023instaflow,dieleman2024distillation}.
%


\noindent
\textbf{Markov Random Fields:}
Before the deep learning era, MRFs played a significant role in solving many vision problems such as segmentation, reconstruction, and denoising~\citep{szeliskimrf}. Recently, fully learnable variants (i.e. ones with trainable parameters that can learn pairwise and higher-order compatibility from data) have been developed. The energy minimization is cast as a neural network layer using mean field approximate inference techniques, and solved through backpropagation~\citep{Zheng2015ConditionalRF,Yang2021ContinuousCR,Jayasumana2023MarkovGenSP,Guan2024NeuralMR}. Such techniques have been used in image generation, where the goal is to predict discrete labels of quantized image tokens~\citep{Jayasumana2023MarkovGenSP}. 

To speed up diffusion, we are interested in denoising latent vectors, and the method of~\citep{Jayasumana2023MarkovGenSP}, which predicts discrete labels is not applicable as it doesn't handle continuous latent vectors. A few methods in other domains have looked at continuous MRFs such as point-cloud segmentation~\cite{Yang2021ContinuousCR,Zheng2015ConditionalRF}, but \modelname is distinctly different in the choice of underlying energy functions, the use of higher order terms, conditioning on text tokens, and the loss functions and training algorithm for text-to-image generation.

\section{Methodology}
\label{sec:method}
In this section, we introduce \modelname for encoding the spatial and semantic relationships among feature vectors in the latent space. We will explain the \modelname formulation with the underlying energy function, and the associated loss functions used to train it as a neural network layer. 



\subsection{Problem Formulation}
We consider the standard LDM setting~\cite{Rombach2021HighResolutionIS} with a pretrained autoencoder with encoder $\gE$ that maps an image in $\sR^{h \times w \times c}$ to latent $\vz \in \sR^{h' \times w' \times d}$ and a decoder $\gD$ that projects $\vz$ back to $\sR^{h \times w \times c}$. 
The LDM model is given by $\gU: \sR^{h' \times w' \times d} \rightarrow \sR^{h' \times w' \times d}$, that operates within the latent space of this autoencoder. The LDM ($\gU$) is trained to map Gaussian noise to latent vectors of natural images through an iterative reverse diffusion process and is conditioned on textual inputs to enable text-to-image generation. A noise vector is iteratively refined to obtain a specific latent vector, which is then mapped to pixel space with the decoder ($\gD$), constructing the final image.

The standard approach for inference (i.e. text-to-image generation) with such an LDM is to iteratively apply $\gU$ conditioned on a textual prompt to denoise the input noise vector, obtaining the desired image at the end of the iterative process. 
Standard MRFs (including in~\cite{Jayasumana2023MarkovGenSP}) operate on discrete probability distributions while LDMs operate in a \textit{continuous} latent space. To address this and handle LDMs, we propose a continuous CRF model to handle both inputs and outputs in continuous space and denote this as \modelname $\gM: \sR^{h' \times w' \times d} \rightarrow \sR^{h' \times w' \times d}$.


\subsection{The \modelname model}
Let $i \in \{1, 2, \dots, n\}$ denote location indices of a latent image $\vz \in \sR^{h' \times w' \times d}$, where $n=h' w'$. We define random variables $Y_i \in \sR^d$ for $i = 1, 2, \dots, n$, to represent the $d$-dimensional latent vector at the $i^\text{th}$ location of $\vz$. The collection of these random variables $\mathbf{Y} = (Y_1, Y_2, \dots, Y_n)$ forms a random field, where the value of each random variable depends on those of others. We can model the probability of an assignment to this random field with the Gibbs measure:
\begin{equation}
    P(\Y = \y) = \frac{1}{Z}\exp(-E(\y)),
    \label{eq:energy_prob}
\end{equation}
where $\y = (\vy_1, \vy_2, \dots, \vy_n)$ is a given assignment to the random field $\Y$. The term $E(\y)$ is known as the \emph{energy} of the assignment $\y$, and $Z$ is a normalizing constant. Let $\x = (\vx_1, \vx_2, \dots, \vx_n)$ be an observed ``noisy" assignment to $\Y$. Our goal is to identify a denoised assignment $\y$ given a noisy observation $\x$. To achieve this, we design $P(\y|\x)$ (equivalently, $E(\y|\x)$) in a way that assigns higher probability (equivalently, lower energy) to desirable assignments. 

We write the energy of the ``denoised" assignment $\y$ as:
{\small%
\begin{align}
    E(\y|\x) = \sum_i d(\vy_i, \vx_i) + \sum_{i,j} f_{ij}(\vy_i, \vy_j, \vc) + \sum_k g(\y_k),%
\end{align}}%
where $d(\vy_i, \vx_i)$  measures the energy component corresponding to each latent $i$ independently, $f_{ij}(\vy_i, \vy_j, \vc)$ measures the energy component arising from pairwise interactions between latent pairs $(i, j)$ conditioned on additional data $\vc$ (such as a text description of the image), and $g(\y_k)$ denotes the energy component due to higher-order interactions in latent cliques $k \subset \{1, ..., n\}$ formed with more than two latents. 
This formulation, using probabilistic graphical models, lets us interpret the exact role of each of the energy terms. We next discuss these terms.

\subsubsection{Unary Energy}
Recall that our latent diffusion model assumes the observed noisy data $\x$ is generated from the true underlying data $\y$ by adding Gaussian noise (see \Cref{app:forward_diffusion} for details).
We therefore have, $P(\y|\x) \propto \exp(-\|\y - \x\|^2)$. Since energy corresponds to negative log-probability (\cref{eq:energy_prob}), this leads to the following unary energy: $ d(\vy_i, \vx_i) = \|\vy_i - \vx_i\|^2$.


This energy can be intuitively understood as penalizing deviations of $\vy_i$ from $\vx_i$.

\subsubsection{Pairwise Energy}
The pairwise energy function $f$ captures the mutual compatibility of latent vector configurations in the distribution of natural images. For example, if a latent vector represents a human eye, then a spatially nearby latent will likely represent other parts of a face. 
This semantic relationship will be modeled by the following pairwise term:
\begin{equation}
    f_{ij}(\vy_i, \vy_j, \vc) = \Wsij\,\| \Wc \vy_i - \Wc \vy_j \|^2,
\end{equation}
where the scalar $\Wsij$ scores the similarity between the latent $i$ and latent $j$ (e.g. nearby latent pairs may have high similarity), and the feature compatibility matrix $\Wc$ 
projects latent vectors to a new space where ``compatible" vectors will be close to each other. 
In text-to-image generation, the notion of realism could be dependent on the text prompt, leading to our conditioning of this energy term on an additional text condition $\vc$ that is incorporated into the feature compatibility matrix. 
Transformations of this nature are referred to as the \emph{compatibility transform} in the CRF literature. 
In our formulation, both $\Ws = [\Wsij]$ and $\Wc$ are learned from data with the latter additionally conditioned on textual inputs. 

\subsubsection{Higher-Order Energy}
\label{sec:higher-order}
For the higher-order energy term, we use a Field-of-Experts (FoE) model~\cite{Roth2005FieldsOE}, where we model the compatibility of a patch (i.e., a higher-order clique of latents) with a collection of filter responses. Let $\y_k$ be a latent patch of shape $h'' \times w''$. It is well known that only a small subset of latent vector arrangements in a given patch represent realistic looking image patches \cite{Hu2024StructLDMSL,Zhang2024MultiScaleDE}. 
It was shown in works such as \cite{Hinton1999ProductsOE, Roth2005FieldsOE, NIPS2002_bb1662b7}, albeit in the domain of natural image pixel-space, that statistics of valid patches can be effectively learned with a bank of filters $\J_1, \J_2, \dots, \J_M$, each having the same dimensions as $\y_k$. We follow a similar Product-of-Experts model~\cite{Hinton1999ProductsOE}, but within latent space and define the probability of a latent image patch $\y_k$ as follows:
\vspace{-0.5em}
\begin{equation}
    P(\y_k) = \frac{1}{Z_k}\prod_{m=1}^M \phi\left(\J_m^T\y_k\right),
    \label{eq:ho_poe}
    \vspace{-0.2em}
\end{equation}
%
where $\phi(.)$ is an appropriate probability density function without the normalization term and $Z_k$ is the partition function that takes care of the normalization of the probability density. We assume the filter response $\J_m^T \y_k$ is bounded. In our inference algorithm, $\y_k$ remains bounded. Therefore, this can be implemented by either normalizing the filter response or normalizing / clipping filter coefficients. 
We further assume that the greater the alignment between a filter and a patch, the larger the likelihood of the patch. Given these, it is preferable for $\phi(.)$ to be a monotonically increasing function. Consequently, the requirements for $\phi(.)$ are that it must be non-negative and monotonically increasing, with normalization addressed by the partition function.  

This gives rise to the following higher-order energy term for one patch $\y_k$:
\vspace{-1.0em}
\begin{equation}
g(\y_k) = -\sum_{m} \log \phi \left(\J_m^T\y_k\right).    
\vspace{-0.5em}
\end{equation}
This is then extended to the whole latent image by considering all $\y_k$ patches centered at each latent $k$:
\vspace{-0.5em}
{\small%
\begin{align}
    \sum_k g(\y_k) &= \sum_{k=1}^n  \sum_{m} -\log \phi (\J_m^T\y_k )\nonumber\\
    &= \sum_{k=1}^n \sum_m \Big[ -\log \phi (\J_m \odot \y )\Big]_k,
    \vspace{-0.5em}
\end{align}}%
where $\J_m \odot \y$ denotes 2D convolution of the latent image $\y$ with the filter $\J_m$ and $[\,.\,]_k$ denotes selecting the $k^{\text{th}}$ latent of the convolution output.

\subsubsection{Overall Energy Function}
The full energy function takes the form:
\vspace{-0.0em}
{\small%
\begin{align}
    E(\y|\x) = \sum_i &\|\vy_i - \vx_i\|^2 + \sum_{i,j} \Wsij\,\| \Wc\vy_i - \Wc\vy_j \|^2 \nonumber\\ 
    &+ \sum_i \sum_m \Big[ -\log \phi (\J_m \odot \y )\Big]_i.
    \label{eq:energy_final_form}
    \vspace{-1.0em}
\end{align}}%
Typically, higher order energy terms can implicitly model pairwise as well, but the long-range interactions in pairwise terms are not covered by the local FoE based-filters in higher order terms that only consider the interaction of spatially nearby latents. The exact form of $\phi(.)$ is chosen to facilitate a straightforward implementation of the inference algorithm, as discussed in next section.


\subsection{\modelname Inference Algorithm}
\label{subsec:mrf_infer}

We now discuss inference with our CRF formulation in \Cref{eq:energy_final_form}. CRF inference finds a denoised configuration $\y$, given a noisy (observed) configuration $\x$, by minimizing the energy $E(\y|\x)$. While different methods exist to minimize Gibbs energy functions, we would like to use a fully differentiable algorithm that would allow us to use CRF inference as a layer in a neural network. To this end, we utilize the differentiable mean-field inference popularized by works such as \cite{Krahenbuhl2011, Zheng2015ConditionalRF, Yang2021ContinuousCR, Jayasumana2023MarkovGenSP}.

In this method, $E(\y|\x)$ in \Cref{eq:energy_final_form} is minimized with coordinate descent by successively minimizing $E$ with respect to $\vy_i$'s. The update at each step can be derived as:
\vspace{-0.0em}
{\small %
\begin{align}
    \frac{\partial E(\y)}{\partial \vy_i} = 2(&\vy_i - \vx_i) + 2 \sum_j \Wsij \Wc^T\Wc (\vy_i - \vy_j) \nonumber \\
    &- \sum_m \Big[  \J_m^{-} \odot \omega(\J_m \odot \y )  \Big]_i= 0,
    \label{eq:delta_energy}
\end{align}}%
where $\omega(y) = \frac{\partial}{\partial y} \log \phi(y)$, $\J_m^{-}$ is the mirrored (left-right and up-down around the center latent) version of the filter $\J_m$.
This yields the following iterative update equation for $\vy_i^{(t+1)}$ given the previous iteration's estimates,  $\y^{(t)}$:
\vspace{-0.5em}
{\small%
\begin{align}
\label{eq:update}
    \vy_i^{(t+1)} := \ \rmK^{-1} \bigg(& \vx_i + \Wc^T\Wc \sum_j \Wsij \vy_j^{(t)} + \nonumber \\ 
    & \frac{1}{2} \sum_m \Big[ \J_m^{-} \odot \omega\big(\J_m \odot \y^{(t)} \big) \Big]_i\bigg), \nonumber \\
    \text{with}\;\; \rmK = \Big(\rmI\,+\,&\Wc^T\Wc \sum_j \Wsij\Big), 
    \vspace{-0.5em}
\end{align}}%
where $\rmI$ is the identity matrix\footnote{We have neglected the contribution of $\vy_i$ in $\J_m \odot \y^{(t)}$ to reduce clutter. This term can be accounted for by absorbing it into $\rmK$, or it can be eliminated by setting the center latent of $\J_m$ to zero.}.

\begin{algorithm}[t]
\caption{\it The \modelname Inference Algorithm}
\label{alg:mrf_infer}
\begin{algorithmic}
\State $\{\vy_1, \vy_2, \dots, \vy_n \} = \y \gets \x$ \Comment{Initialization}
\State $\vc \gets \texttt{text\_embedding} $  \Comment{Text Condition}
\vspace{0.2em}
\For {\texttt{num\_iterations}}
    \State $\check{\vy}_i \gets \sum_j \Wsij\, \vy_j$, $\forall i$ \Comment{Message Passing}
    \State $\grave{\y} \gets \psi_{\theta_C}(\check{\y}, \vc)$ \Comment{Compatibility Transform}
    \State $\breve{\y} \gets \grave{\y}  + \xi_{HO}(\y) $ \Comment{Higher-Order}
    \State $\Tilde{\y} \gets \breve{\y} + \x$ \Comment{Unary Addition}
    \State $\y \gets \psi_{N}(\Tilde{\y})$ \Comment{Normalization}
\EndFor
\vspace{0.2em}
\State \textbf{return} $\y$
\end{algorithmic}
\end{algorithm}
\begin{figure*}[t]
    \begin{minipage}{\linewidth}
        \centering\centering
        \includegraphics[width=\linewidth]{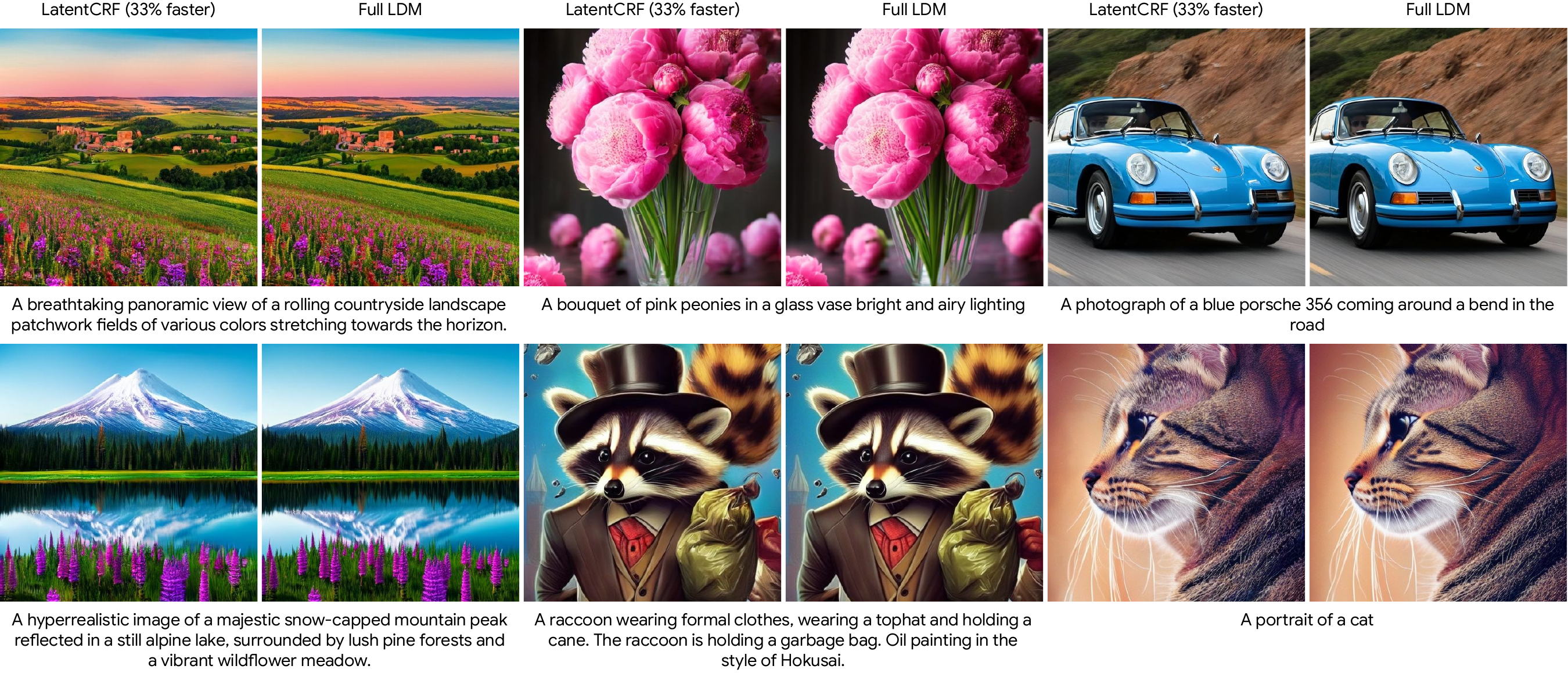}
    \end{minipage}
    \vspace{-0.5em}
    \caption{
    \textbf{Qualitative Results:} Within each set of two, LatentCRF (left) speeds up LDM (right) by 33\% while maintaining image quality.}
    \label{fig:qualitative_results}
    \vspace{1.0em}
\end{figure*}

%




We make several modifications to the update in \Cref{eq:update} with the aim of a richer CRF model with stable and efficient training and inference properties. First, we employ parallel updates instead of sequential updates. While the usual convergence proofs for mean-field inference do not hold for parallel updates, in practice, fast convergence is observed within a few iterations (see \cite{Krahenbuhl2011} and \Cref{app:mrf_converge}). 
Next, for the pairwise term, we replace the compatibility transform with a lightweight nonlinear neural network $\psi_{\theta_C}$ with the aim to capture more complex compatibility relationships between latent vectors. We also replace the normalization term with a learnable normalizing operation $\psi_{N}$ (i.e. batch-norm) for training stability. 

%
For the term $\phi(.)$, as discussed in \Cref{sec:higher-order}, the requirements are that it must be nonnegative and monotonically increasing. We use the following function, which has both these properties and the additional desirable property that $\frac{\partial}{\partial y} \log \phi(y) = \omega(y) =  \operatorname{ReLU}(y)$ (see \Cref{app:foe_details} for further details), which is a widely utilized function in deep learning:
\begin{equation}
    \phi(y) = \begin{cases}
                        e^{y^2/2},\;\;\;\; y > 0 \\
                        \varepsilon,\;\;\;\;\;\;\;\; \text{otherwise,}
              \end{cases}
\end{equation}
where $\varepsilon > 0$ is a small constant. 

Therein we introduce a modified version of \cref{eq:update} as,
{\small%
\begin{align}
\label{eq:final_update}
    \vy&_i^{(t+1)} := \psi_{N} \bigg(\vx_i + \psi_{\theta_C} \Big( \sum_j \Wsij \vy_j^{(t)} \Big) +  \xi_{HO}(\y^{(t)}) \bigg), \nonumber \\
    &\text{with}\;\; \xi_{HO}(\y^{(t)}) = \frac{1}{2} \sum_m \Big[ \J_m^{-} \odot \omega( \J_m \odot \y^{(t)} ) \Big]_i . 
\end{align}}%
Following \Cref{eq:final_update}, we present our complete CRF inference algorithm in \Cref{alg:mrf_infer}. It is straightforward to efficiently implement in any deep learning framework thanks to the usage of ubiquitous operations such as matrix multiplication, convolution, and the $\operatorname{ReLU}$ activation function. We refer to the application of this update rule, i.e. inference of our CRF module, as a function denoted by $\gM$. $\texttt{num\_iterations}$ is a hyper-parameter which we set at 5, it can be tuned for balance between convergence and run time. Further details on setting this value can be found in \Cref{app:num_iterations}.

We refer the reader to our appendix for further implementation details and turn to the training procedure for our CRF module.

\subsection{LatentCRF model training}\label{sec:natural_image_training}
To teach LatentCRF about the distribution of natural images within a continuous latent space, we propose a training with two loss terms, a denoising loss and an adversarial loss. 

\subsubsection{Latent Denoising Loss} 
Given a natural image and corresponding latent $\vz$, we create a noisy version
$\tilde{\vz} = \sqrt{\alpha} \cdot \vz + \sqrt{1 - \alpha} \cdot \vn$, where $\vn \sim \mathcal{N}(0, 1)$ is noise drawn from a normal distribution and $\alpha$ is a hyper-parameter controlling the noise ratio. Our denoising loss aims to minimize the distance between the denoised CRF output and the clean latent as $\gL_{\text{NT}} = \|\vz - \gM(\tilde{\vz})\|_2$.




\subsubsection{Latent Adversarial Loss} 
In contrast to pixel space, slight perturbations in our continuous latent space sometimes result in no visible change in pixel-space (see \Cref{app:ladv} for details). Therefore, training the CRF using only $\gL_{\text{NT}}$ may be sub-optimal (e.g. we sometimes observe visual artifacts and slightly blurrier outputs). Furthermore, our latent space contains spatial dimensions encoding structure that adversarial losses are known to capture better \cite{goodfellow2014generative}. 

Motivated by these observations, we introduce a latent space adversarial objective, $\gL_{\text{NT-ADV}}$ where a secondary discriminator neural network operates on the reconstructed latents, $\gM(\tilde{\vz})$. We construct a simple 3 layer convolutional neural network (more details in \Cref{app:ladv}) with a series of discriminator heads motivated by \cite{Sauer2023StyleGANTUT,Sauer2023AdversarialDD}, but in contrast operating directly in latent space. 
This discriminator, $\theta_{D}$, is applied on the generated and original (natural image) latents to calculate a non-saturating sigmoid cross-entropy loss as,
{\small%
\begin{equation}
    \gL_{\text{SCE}}(a, t) =  \mathtt{max}(a, 0) - a \cdot t + \mathtt{log}(1 + \mathtt{exp}(-\mathtt{abs}(a))) ,
\end{equation}}%
where $a$ is the discriminator output and $t \in \{0, 1\}$, for generated or original latents respectively. 

Our combined loss for training the CRF ($\gL_{\text{NT-F}}$) and the discriminator loss ($\gL_{\text{disc}}$) is as follows, 
\begin{align}
    \gL_{\text{NT-F}} &=  \gL_{\text{NT}} + \gL_{\text{SCE}}(\gM(\tilde{\vz}), 1) \\
    \gL_{\text{disc}} &=  \gL_{\text{SCE}}(\vz, 1) + \gL_{\text{SCE}}(\gM(\tilde{\vz}), 0).
    \label{eq:natural_image_loss}
\end{align}

\begin{table}[t]
\small
\centering
\caption{
\textbf{Quantitative Evaluation of \modelname:} 
We measure quality with FID and CLIP scores (on 30K COCO images), diversity with Vendi scores, and speed in ms per image generation. `Iter' refers to the number of LDM U-Net iterations used to sample a single image by each row. Both variants of our proposed \modelname framework improve image quality at a higher inference speed, with no or minimal diversity losses.
}
\label{table:results_main}
\vspace{-1.0em}
\def\arraystretch{1.2}  
\setlength\tabcolsep{0.6em}  
\scalebox{0.88}{
\begin{tabular}{l|c|c|c|c|c}
\toprule
Method   & Iter & FID $\downarrow$& CLIP $\uparrow$& Vendi $\uparrow$ & \text{Speed (ms)} $\downarrow$ \\ \midrule
LDM \cite{Rombach2021HighResolutionIS}  &  50  & 12.75  & 0.309  & 2.75  & 782.1  \\
LDM \cite{Rombach2021HighResolutionIS}  &  33  & 12.78  & 0.308  & 2.57  & 501.8\inc{35.9\%}  \\ \rowcolor{Gray}
\modelname                              &  33  & 11.58  & 0.309  & 2.64  & 523.4\inc{33.2\%}  \\ \midrule
LDM-L \cite{Rombach2021HighResolutionIS}&  50  & 12.23  & 0.312  & 2.82  & 2335       \\
LDM-L \cite{Rombach2021HighResolutionIS}&  33  & 12.35  & 0.311  & 2.80  & 1528\inc{34.6\%} \\ \rowcolor{Gray}
\modelname-L                            &  33  & 11.27  & 0.311    & 2.82  & 1549\inc{33.7\%} \\ 
\bottomrule
\end{tabular}}
\end{table}

\begin{table}[t]
\small
\centering
\caption{
\textbf{Prior Work Comparison:} 
We compare against prior works that use model compression or temporal distillation to speed up LDM inference, and demonstrate the competitive image quality of our proposed \modelname.
}
\label{table:res_compress}
\vspace{-1.0em}
\def\arraystretch{1.1}  
\setlength\tabcolsep{1.3em}  
\scalebox{0.90}{
\begin{tabular}{lcc} 
\toprule
   Model & FID $\downarrow$ & CLIP $\uparrow$ \\ \midrule
    SnapFusion~\cite{li2023snapfusion}           & 14.00 & 0.300 \\
    BK-SDM-Base \cite{kim2023architectural}      & 17.23 & 0.287 \\
    BK-SDM-Small \cite{kim2023architectural}     & 17.72 & 0.268 \\
    BK-SDM-Tiny \cite{kim2023architectural}      & 18.64 & 0.265 \\
    InstaFlow (1 step) \cite{liu2023instaflow}   & 20.00 & 0.283 \\
    Clockwork \cite{Habibian2023ClockworkDE}     & 12.33 & 0.296 \\ 
    LCM (8 steps) \cite{Luo2023LatentCM}         & \underline{11.84} & 0.288 \\
    ADD (4 steps) \cite{Sauer2023AdversarialDD}  & 22.58 & \textbf{0.312} \\     \rowcolor{Gray}
    \modelname (ours)                            & \textbf{11.58} & \underline{0.309} \\
\bottomrule
\end{tabular}
}
\end{table}


\begin{table}[t]
\small
\centering
\caption{
\textbf{Diversity Evaluation:} 
We report Vendi Score and its percentage drop relative to teacher model and compare against state-of-the-art temporal distillation method, ADD \cite{Sauer2023AdversarialDD}. \textit{Steps} denotes the number of U-Net iterations. $^\dagger$ indicates teacher model.
}\label{table:diversity}
\vspace{-1.0em}
\def\arraystretch{1.2}  
\setlength\tabcolsep{1.3em}  
\scalebox{0.85}{
\begin{tabular}{l|c|S}
\toprule
Method  & Steps & \text{Vendi Score $\uparrow$} \\ \midrule
SDXL \cite{Podell2023SDXLIL} $^\dagger$       &  50  & 3.01  \\ 
ADD  \cite{Sauer2023AdversarialDD}  &   1  & 1.55\dec{48.5\%}  \\ 
ADD  \cite{Sauer2023AdversarialDD}  &   4  & 1.69\dec{43.9\%}  \\ 
ADD  \cite{Sauer2023AdversarialDD}  &  33  & 2.02\dec{32.9\%}  \\ 
ADD  \cite{Sauer2023AdversarialDD}  &  50  & 2.05\dec{31.9\%}  \\ \midrule
LDM \cite{Rombach2021HighResolutionIS} $^\dagger$ &  50  & 2.75  \\ \rowcolor{Gray}
\modelname (ours)                   &  33  & 2.64\dec{4.00\%}  \\ 
LDM-L \cite{Rombach2021HighResolutionIS} $^\dagger$ &  50  & 2.82  \\ \rowcolor{Gray}
\modelname-L (ours)                   &  33  & 2.82\inc{0.00\%}  \\ 
\bottomrule
\end{tabular}}
\end{table}
\begin{table*}
\small
\centering

\begin{minipage}{0.32\textwidth}
\centering
\caption{\footnotesize
\textbf{Inference Pipeline Ablation:} 
We demonstrate effects of ablating (one or both of) the `CRF' and final 2 `Post-CRF' LDM iterations from our inference pipeline. Both contribute to image quality. (The LDM baseline is reported for comparison purposes.
}\label{table:ablate_infer}
\vspace{-1.0em}
\def\arraystretch{1.2}  
\setlength\tabcolsep{0.5em}  
\scalebox{0.78}{
\begin{tabular}{c|c|c|c|l}
\toprule
Method  & Pre-CRF & CRF    & Post-CRF & \text{FID $\downarrow$} \\ \midrule \rowcolor{Gray}
Ours    & 31      & \cmark &  2       & 11.58 \\
Ours    & 31      & \cmark &  0       & 11.74\dec{0.16} \\
Ours    & 31      & \xmark &  2       & 19.72\dec{8.14} \\
Ours    & 31      & \xmark &  0       & 21.85\dec{10.27} \\ 
\midrule
LDM  & -      & -      &  -      & 12.75 \\
\bottomrule
\end{tabular}
}
\end{minipage}
%
\hspace{0.01\textwidth}
\begin{minipage}{0.32\textwidth}
\centering
\caption{\footnotesize
\textbf{CRF Ablation:} 
We report FID on base (FID-B) and large (FID-L) variants of \modelname where CRF sub-components are ablated. Removing the pairwise (only unaries) will provide inputs directly as outputs, so it is omitted. Adding text conditioning (text) to the pairwise term and introducing the higher order term, consistently improves FID across both variants.   
}\label{table:ablate_energy}
\vspace{-1.0em}
\def\arraystretch{1.2}  
\setlength\tabcolsep{0.5em}  
\scalebox{0.84}{
\begin{tabular}{c|c|l|l}
\toprule
Text  & Higher Order & \text{FID-B $\downarrow$} & \text{FID-L $\downarrow$} \\ \midrule \rowcolor{Gray}
\cmark & \cmark &  11.58           & 11.27 \\
\cmark & \xmark &  11.62\dec{0.04} & 11.58\dec{0.31} \\ 
\xmark & \xmark &  11.78\dec{0.20} & 11.72\dec{0.45} \\
\bottomrule
\end{tabular}
}
\end{minipage}
%
\hspace{0.01\textwidth}
\begin{minipage}{0.32\textwidth}
\caption{\footnotesize
\textbf{LDM Schedule in \modelname Pipeline:} 
We explore integrating sparser LDM schedules with \modelname for further speed-up and uncover promising results but with some image quality drop. More details in \Cref{app:sparse_schedules}. 
}
\label{table:ablate_schedule}
\vspace{-1.0em}
\def\arraystretch{1.2}  
\setlength\tabcolsep{0.8em}  
\scalebox{0.85}{
\begin{tabular}{l|c|c|c}
\toprule
Method   & Iter & FID $\downarrow$ & \text{Speed (ms)} $\downarrow$ \\ \midrule
LDM \cite{Rombach2021HighResolutionIS}  &  50  & 12.75  & 782.1  \\
LDM \cite{Rombach2021HighResolutionIS}  &  33  & 12.78  & 501.8\inc{35.9\%}  \\
LDM \cite{Rombach2021HighResolutionIS}  &  20  & 12.80  & 323.9\inc{58.6\%}  \\ \midrule 
Ours                              &  33  & 11.58  & 523.4\inc{33.2\%}  \\ 
Ours                             &  20  & 11.67  & 344.7\inc{55.9\%}  \\ 
\bottomrule
\end{tabular}}
\end{minipage}
%

\end{table*}

\subsection{Distillation with LDM}\label{sec:distillation_with_ldm}
Our main goal is to swap the computationally intensive LDM steps with lightweight \modelname inference. In order to do this, we distill from LDM in such a manner that \modelname can mimic some of the inference steps of LDM.

Given a dataset of text prompts (or captions describing natural images), we use the LDM to iteratively generate corresponding images. For some intermediate iteration, $s$, we extract the latent ${\vz}_s$ output by the LDM U-Net at that step. We also extract the final latent, ${\vz}_f$, corresponding to the final generated image. Therein, we introduce our distillation training objective: $
    \gL_{\text{DT}} = \|{\vz}_f - \gM({\vz}_s)\|_2$.
We initialize distillation training with the checkpoint from the first stage. 

\section{Experiments}\label{sec:experiments}
\label{sec:exp}


\noindent \textbf{Training Setup:}
Our VAE and LDM are trained following \cite{Rombach2021HighResolutionIS}, using web-scale image data with paired captions.
The \textit{base} LDM variant has 865 million parameters and the \textit{large} 2 billion (in contrast our CRF contains only 130 million). Both variants generate 512x512x3 RGB images using a 64x64x8 latent space of the VAE. 

Our \modelname is first trained on natural images, as described in \Cref{sec:natural_image_training} for 300,000 iterations with an AdamW optimizer, cosine-decay learning rate, initialized at $1e-3$ with 5000 warm-up iterations. Our per-device batch size is 16 with 128 TPUv5e devices. 
%

Distillation, as described in Section~\ref{sec:distillation_with_ldm}, is initialized from the naural image checkpoint and run for 10,000 iterations with the same optimizers and learning rate scheduler, but with an initial learning rate of $1e-5$ and a per-device batch size of 4. Training uses captions to generate intermediate (step 40 of 50) and final (step 50 of 50) latents.

\noindent \textbf{Inference Pipeline:}
Our inference schedule for \modelname was manually tuned to balance speed and quality. We find that the best results are obtained by replacing later LDM iterations with our CRF. Namely, we run 31 LDM iterations on a 40 step schedule (to accelerate coarse image formation), apply our CRF, and conclude with 2 more LDM iterations (on a 50 step schedule). Initial LDM iterations focus on text-conditioned high variance  changes~\cite{Habibian2023ClockworkDE,Pu2024EfficientDT,Tang2024AdaDiffAD}, where large capacity models play a crucial role. Subsequent LDM iterations focus on photo realism which our lightweight CRF with stronger inductive bias successfully replaces. 
The final 2 LDM iterations touch up fine details.

\noindent \textbf{Evaluation:}
We evaluate \modelname for generation quality, speed, and diversity. 
For quality, we follow \cite{Jayasumana2023MarkovGenSP,Habibian2023ClockworkDE} calculating FID and CLIP scores on 30,000 validation set images from the MS-COCO 2014 dataset.
For speed, we report model inference time, and for diversity, we chose the Vendi score~\cite{Friedman2022TheVS, Burgert2023DiffusionIH}. We calculate this score on $1632 \times 16$ images generated using the 1632 Parti Prompts \cite{yu2022scaling} with 16 different initial noise vectors per prompt (details in \Cref{app:diversity}).

\subsection{Quantitative Evaluations}
\Cref{table:results_main} compares \modelname to (our implementation) of \cite{Rombach2021HighResolutionIS}. During inference, for baselines and \modelname, each LDM iteration is applied with classifier-free guidance with a scale of 7.5 and DDIM~\cite{Song2020DenoisingDI} as the scheduler.
%
Our \modelname outperforms the same-step baseline and virtually matches (or exceeds) the full baseline in terms of both image quality (FID, CLIP score) and diversity (Vendi score). The runs reported in \Cref{table:results_main} use identical settings, including random seeds, as illustrated in \Cref{fig:qualitative_results,fig:diversity}.

We also compare \modelname against prior work that improves LDM inference speed~\cite{Habibian2023ClockworkDE,Luo2023LatentCM}, using their common evaluation settings. The results in \Cref{table:res_compress} show that our \modelname matches or outperforms them in image quality.

We and others~\cite{Sauer2023AdversarialDD} have noted that temporal distillation techniques tend to lose generation diversity. To evaluate this, \Cref{table:diversity} compares \modelname to a state-of-the-art temporal distillation method, ADD~\cite{Sauer2023AdversarialDD}. Comparing Vendi scores of ADD and its teacher SDXL~\cite{Podell2023SDXLIL}, we see diversity drops of over 40\%. In contrast, our \modelname loses only 4\% diversity and \modelname-L loses none, compared to their respective teachers, while achieving a 33\% inference speedup.

It is interesting to note that ADD loses a significant portion of the original teacher model's diversity irrespective of the step count. Even expensive variants using 33 or 50 LDM iterations lose more than 30\% of the teacher's diversity.


\subsection{Ablation Studies}
We ablate our inference pipeline and components of the CRF in \Cref{table:ablate_infer,table:ablate_energy} respectively. The importance of each stage in our inference pipeline is shown in the results in \Cref{table:ablate_infer}. 
\Cref{table:ablate_energy} ablates individual CRF components. Comparing rows 1 and 2 shows the contribution of our higher-order energy term and comparing rows 2 and 3 the importance of text conditioning.

We also explore the effect of LDM schedule sparsity in our inference pipeline. That is, can we replace even more of the LDM iterations? The results presented in \Cref{table:ablate_schedule} are encouraging, they show that \modelname can be integrated with sparser LDM schedules, achieving speed improvements of more then 50\%, but at reduced image quality.

\subsection{Qualitative Evaluations}
We illustrate how our \modelname matches the visual quality of the baseline LDM \cite{Rombach2021HighResolutionIS} in \Cref{fig:qualitative_results}. Careful examination of our results reveals slight color changes and blurred out artifacts which could explain some of the improvements in FID scores we see in~\Cref{table:results_main}.

Diversity in generated images, keeping the prompt fixed and varying the initial noise is illustrated in \Cref{fig:diversity}. Our \modelname retains almost all of the diversity of the LDM teacher. In contrast, ADD~\cite{Sauer2023AdversarialDD} shows decreased diversity (as quantified in~\Cref{table:diversity}). 

\section{Discussion}





\noindent
{\bf Limitations:} Even though our method achieves better FID than the LDM model, we observe some loss patterns where \modelname produces artifacts, such as breaking lines in man-made structures, as shown in \Cref{app:qualitative}. 

\noindent{\bf Inductive bias:} By carefully analyzing LDM inference iterations, we identified an opportunity to replace costly LDM iterations with a lightweight CRF module. The inductive biases which we baked into the CRF help it learn to mimic multiple iterations of the LDMs U-Net, using an order of magnitude fewer parameters.

\noindent{\bf Summary:} We propose \modelname to learn spatial and semantic relationships among the latent feature vectors used for image generation. By replacing LDM iterations with \modelname we achieve a speedup of 33\% with virtually no losses in image quality or diversity.

\section*{Acknowledgements}
We would like to thank Saurabh Saxena, Jonathon Shlens, Sander Dieleman, and Shlomi Fruchter for technical discussion and support.








{
    \small
    \bibliographystyle{ieeenat_fullname}
    \bibliography{main}
}

\clearpage
\newpage
\setcounter{page}{1}

\maketitlesupplementary

\appendix

\renewcommand{\thetable}{A.\arabic{table}}
\renewcommand{\thefigure}{A.\arabic{figure}}
\setcounter{table}{0}
\setcounter{figure}{0}

\section{LDM inference visualization}
\label{app:ldm_vis}
\begin{figure*}[t]
    \begin{minipage}{\linewidth}
        \centering\centering
        \includegraphics[width=\linewidth]{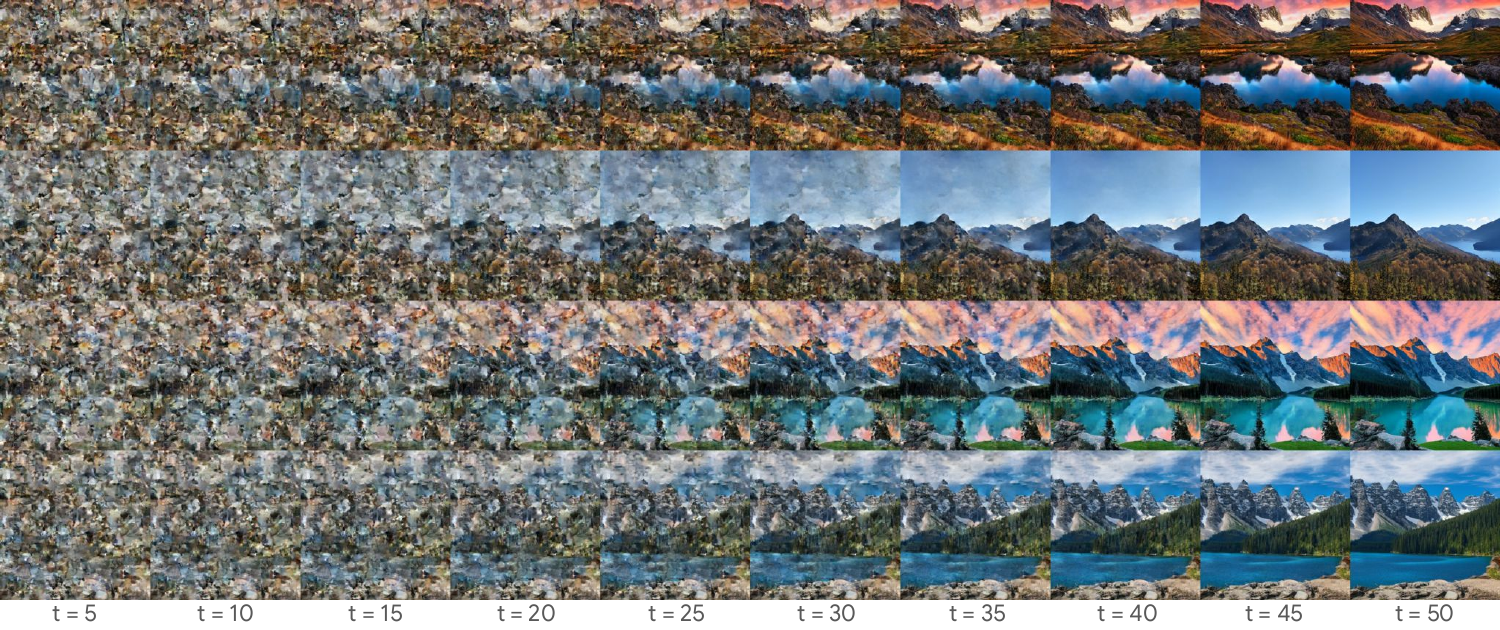}
    \end{minipage}
    \vspace{-0.5em}
    \caption{
    \textbf{LDM Inference:}
    We visualize the reverse diffusion process of LDM, generating images for the prompt
    \texttt{"A beautiful view of a mountain and a lake."} using 50 iterations in a 50 time-step DDIM schedule (default LDM setting used throughout our experiments). Our \modelname focuses on replacing iterations between time-steps 40 to 50 with lightweight CRF modules instead of using the larger LDM U-Net model. As illustrated, almost all text-conditioned elements and structure within an image are formed and fixed by iteration 40. These latter few iterations focus on photo-realism, which we argue does not require a large capacity LDN, unlike the early iterations. Network architectures such as our \modelname utilizing strong inductive biases surrounding natural image structure are more suited for making photo-realism related image modifications. 
    }
    \label{fig:app_ldm_infer}
\end{figure*}

LDM inference involves a reverse diffusion process corresponding to a series of iterative score function evaluations (calls to the LDM U-Net) \cite{Rombach2021HighResolutionIS}. We explore this reverse diffusion process focusing on how latents evolve at each iteration.
Our findings indicate that most early iterations introduce the text-condition-related structure in the image while the latter iterations focus on improving image fidelity. We illustrate this in \Cref{fig:app_ldm_infer}. Notice how by iteration 40 most of the global structure resulting from the text condition (e.g. location and boundaries of the mountains, the lake, the sky) are already fixed. The remaining iterations mainly focus on enhancing the fidelity or photo-realism. 
Similar findings are also discussed in prior work \cite{Habibian2023ClockworkDE,Wimbauer2023CacheMI,si2023freeu} and empirically validated through experimentation in \cite{Pu2024EfficientDT,Tang2024AdaDiffAD}. 

\begin{figure}[h]
    \begin{minipage}{\linewidth}
        \centering\centering
        \includegraphics[width=0.95\linewidth]{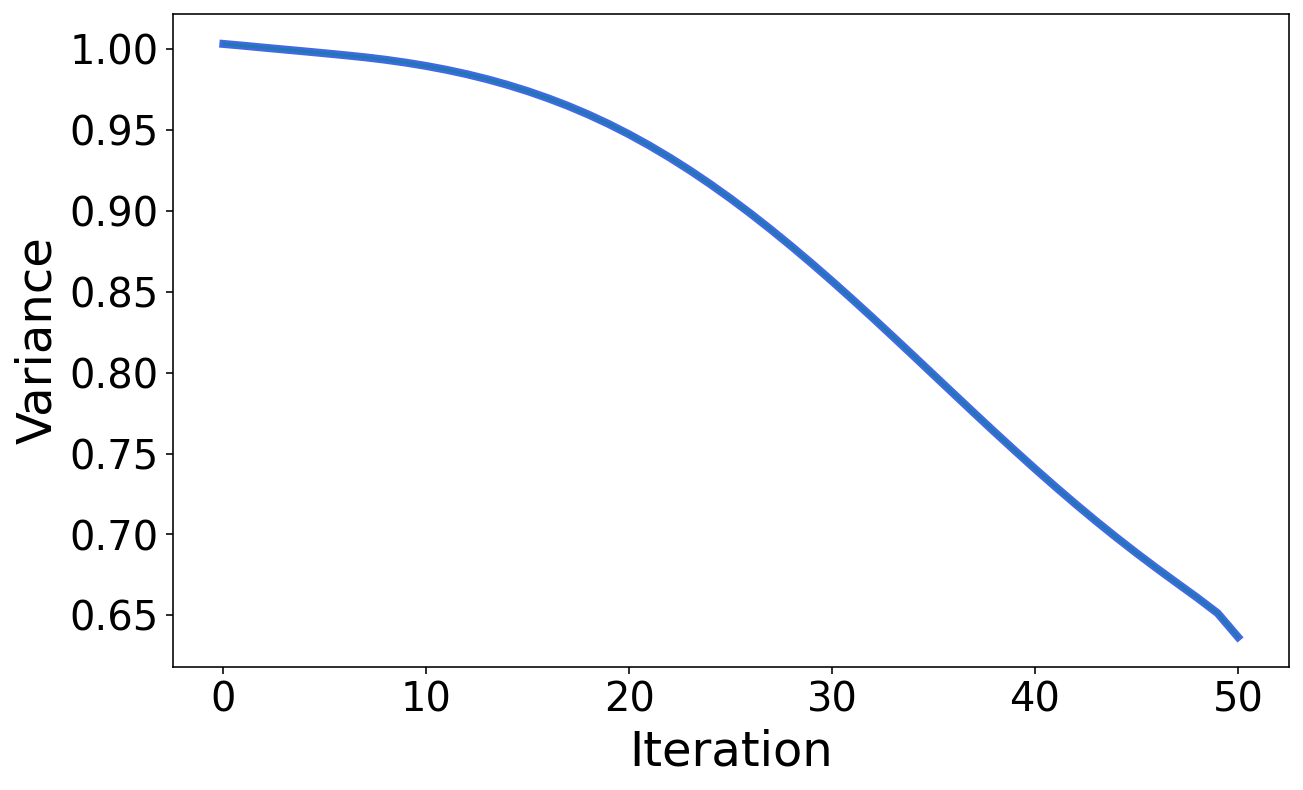}
    \end{minipage}
    \vspace{-0.8em}
    \caption{
    \textbf{Reverse Diffusion Variance:}
    For a 50 time-step DDIM reverse diffusion process, at each iteration we calculate the variance of the generated latents (shaped $h' \times w' \times d$) across all its dimensions. We average these values across 1632 images generated using the Parti Prompts \cite{yu2022scaling}. Note how the variance drops significantly in the latter iterations ($t > 40$), while the early iterations ($t < 20$) in particular exhibit high variance. We intuit that the large capacity U-Nets within an LDM are well suited for guiding latent modifications in this high variance regime. On the other hand, more lightweight modules such as our \modelname are capable of operating in the later, low variance iterations, leading to significant inference speed-ups. 
    }
    \label{fig:app_ldm_variance}
\end{figure}
\begin{figure}[t]
    \begin{minipage}{\linewidth}
        \centering\centering
        \includegraphics[width=\linewidth]{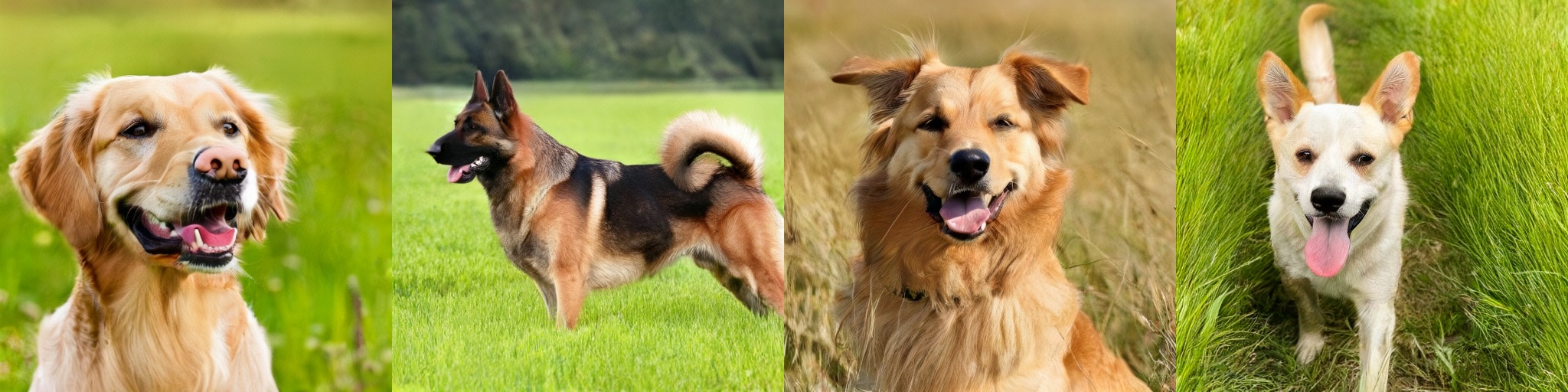}
        \includegraphics[width=\linewidth]{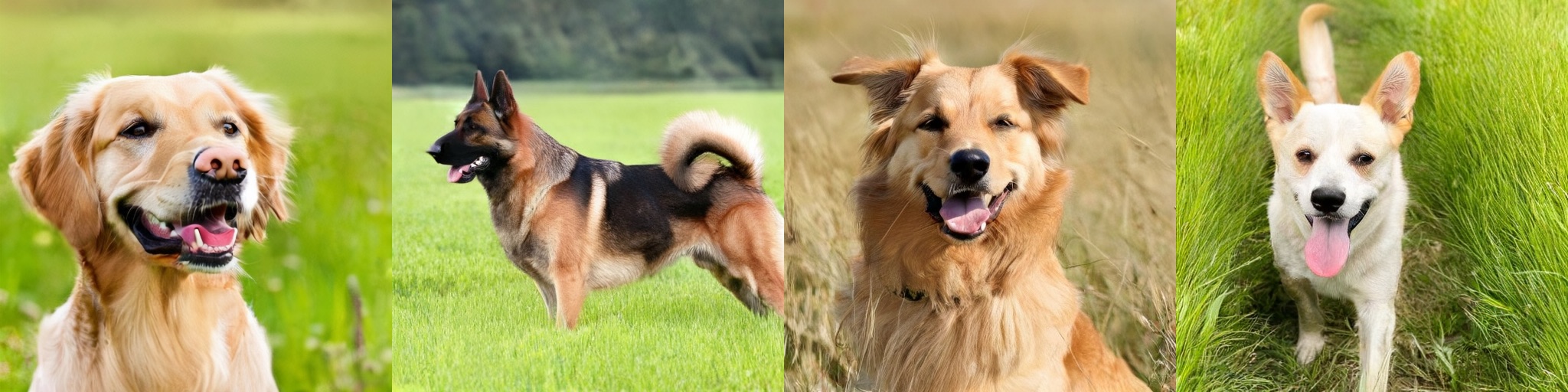}
        \includegraphics[width=\linewidth]{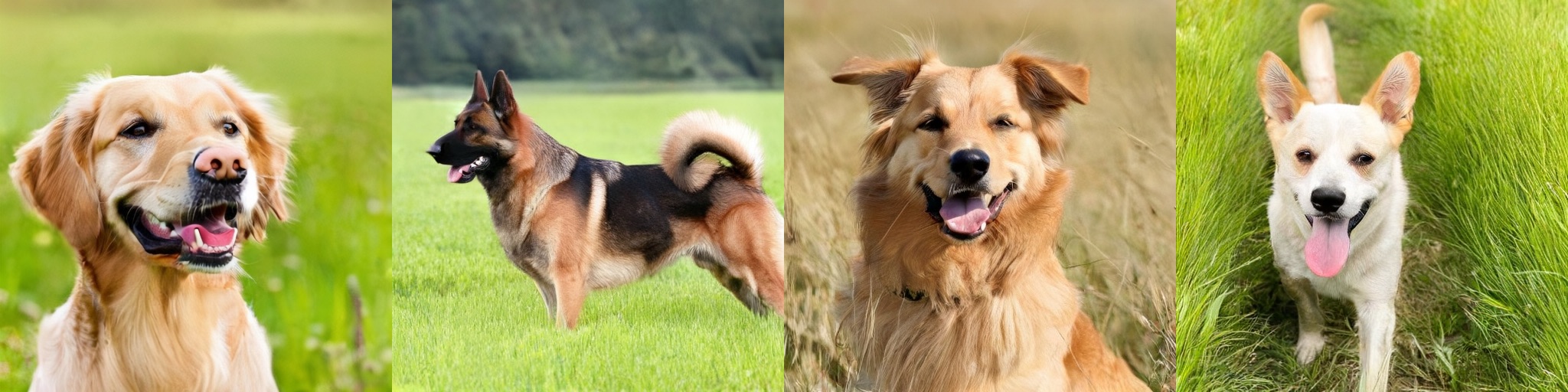}
        \includegraphics[width=\linewidth]{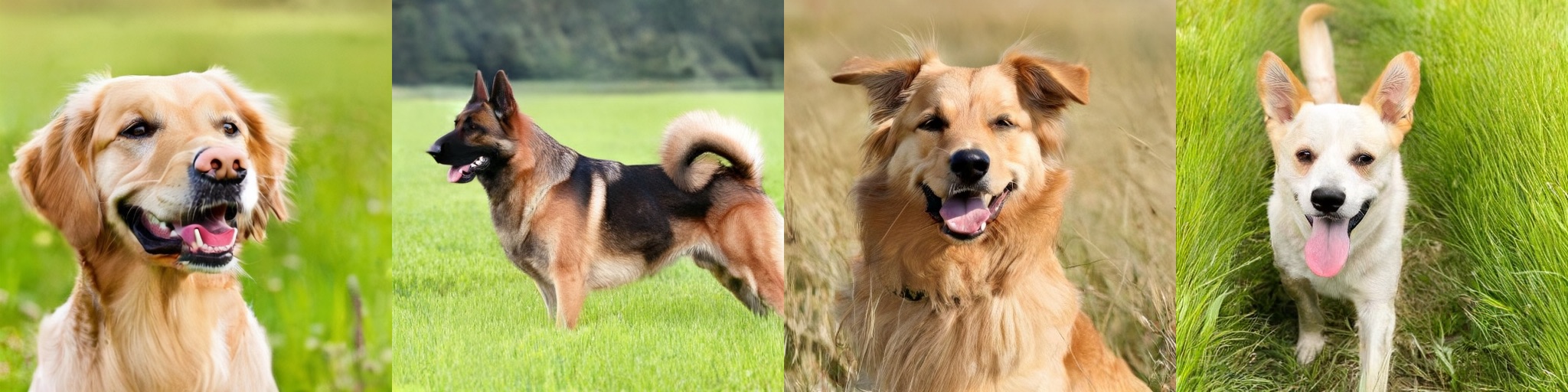}
        \includegraphics[width=\linewidth]{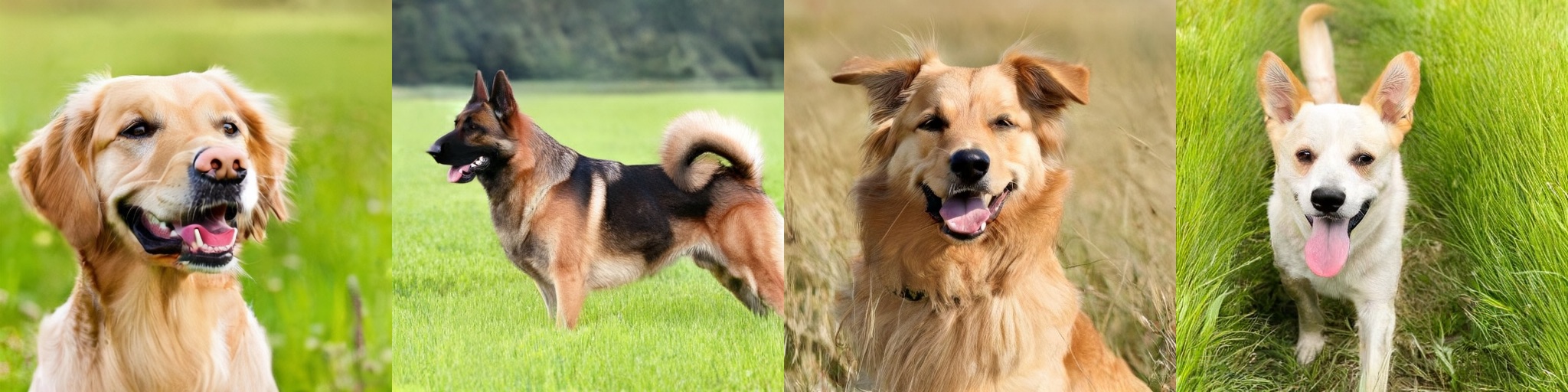}
        \includegraphics[width=\linewidth]{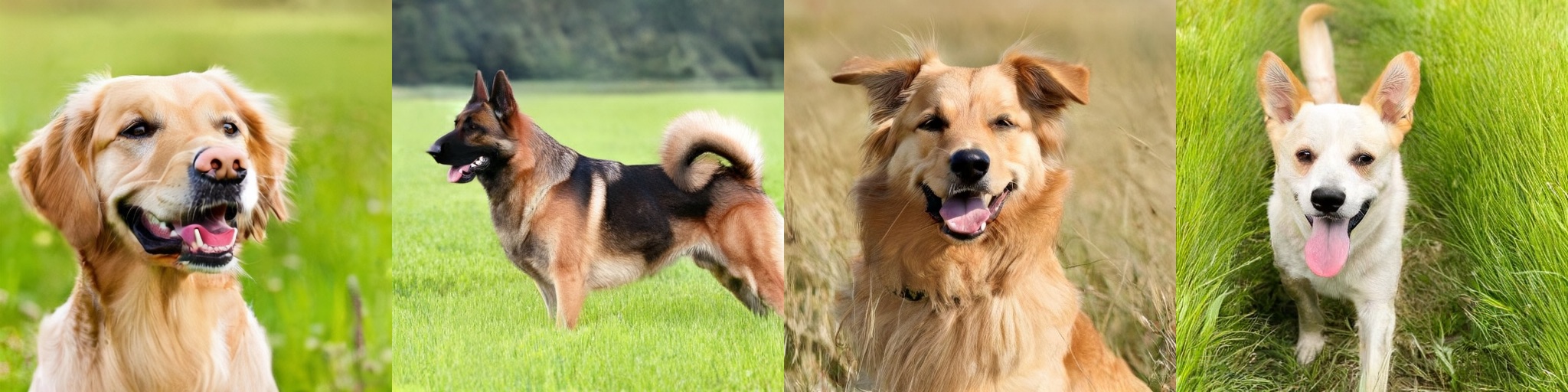}
    \end{minipage}
    \vspace{-0.5em}
    \caption{
    \textbf{CRF Convergence:}
    We visualize images generated by \modelname for varying \texttt{num\_iterations} parameters. We illustrate generated images for values of 1, 2, 3, 4, 5, and 10 in each row from top to bottom respectively. On careful inspection, slight changes are visible in the early iterations (e.g going from iteration 1 to 2 shows color and brightness changes in background). However, the later iterations show no visual changes at all (e.g. beyond step 4), indicating that our CRF inference has converged. All images are generated using the common prompt of \texttt{"A photograph of a dog in a field."}.  
    }
    \label{fig:app_ldm_converge}
\end{figure}

Furthermore, we investigate the variance of latents during the reverse diffusion process for a large set of image generations. We discover how the evolution of latents over time follows a specific pattern and illustrate this in \Cref{fig:app_ldm_variance}. Here we plot the per-sample variance of latents (average over multiple samples) at different iterations.
Latents exhibit high variance during early iterations but lower variance during the latter ones. High variance corresponds to higher entropy within the latents. At the highest (i.e. variance close to 1), the latents correspond to pure noise (see timestep 5 in \Cref{fig:app_ldm_infer}). As the variance reduces, the latents start representing images containing more structure. 
We hypothesize that guiding latents through a high variance (i.e. greater entropy) regime can benefit more from large capacity models such as the LDM U-Net given their stronger representation capacity and stronger textual conditioning. In contrast, we apply our CRF in the lower variance, latter iterations where latents already resemble natural images. This lower variance and closeness to natural images motivates our design to employ a lightweight, lower capacity network with natural image structure tied strong inductive biases in its architecture.

\section{CRF Convergence}
\label{app:mrf_converge}
In \Cref{fig:app_ldm_converge}, we visualize images generated with \modelname using varying values for the \texttt{num\_iterations} parameter in the meanfield inference algorithm (see \Cref{alg:mrf_infer} for details). This variable controls the convergence of MRF inference as discussed in prior work \cite{Krahenbuhl2011,Zheng2015ConditionalRF}. In this example, our proposed \modelname clearly converges by 5 iterations, making no further modifications to the image with additional repeated processing by the network. This behaviour is observed across our evaluation data, where the summed pixelwise difference between generated images at iterations 5 and 10 are almost zero for all samples.

\section{FoE update rule derivation}
\label{app:foe_details}

In \Cref{sec:higher-order} we define higher-order terms for our energy function following FoE \cite{Roth2005FieldsOE} followed by derivation of a suitable inference algorithm in \Cref{subsec:mrf_infer}. We select a suitable probability density function $\phi(y)$ such that $\frac{\partial}{\partial y} \log \phi(y) = \omega(y) =  \operatorname{ReLU}(y)$ to provide a simple and training-efficient implementation while adhering to the non-negative and monotonically increasing requirement. 

In particular, we make the following choice for $\phi(x)$ with some small constant $\varepsilon > 0$:
\begin{equation}
    \phi(y) = \begin{cases}
                        e^{\frac{y^2}{2}},\;\;\;\; y > 0 \\
                        \varepsilon,\;\;\;\;\;\;\;\; \text{otherwise.}
              \end{cases}.
\end{equation}
Taking the derivatives we have,
\begin{equation}
\psi(y) = \frac{\partial \log\phi(y)}{\partial y} = \begin{cases}
                        y,\;\;\;\; y > 0 \\
                        0,\;\;\;\; \text{otherwise.}
              \end{cases}.
\end{equation}
That is, $\psi(x)$ is equal to the $\operatorname{ReLU}$ function commonly used in deep learing. This modification results in our inference algorithm presented as \Cref{eq:final_update}, repeated here as,
{\small%
\begin{align}
\label{eq:final_update}
    \vy&_i^{(t+1)} := \psi_{N} \bigg(\vx_i + \psi_{\theta_C} \Big( \sum_j \Wsij \vy_j^{(t)} \Big) +  \xi_{HO}(\y^{(t)}) \bigg), \nonumber \\
    &\text{with}\;\; \xi_{HO}(\y^{(t)}) = \frac{1}{2} \sum_m \Big[ \J_m^{-} \odot \omega( \J_m \odot \y^{(t)} ) \Big]_i . \nonumber 
\end{align}}%
It is straightforward to implement \Cref{eq:final_update} in a deep learning framework such as JAX with the caveat that the second convolution uses a mirrored version of the first convolution's set of kernels. The filters $\J_i$s can be learned via backpropagation as validated in our experiments. This component alone would serve as a higher order MRF model in the continuous domain since it models the likelihood of patches which are formed with cliques of multiple pixels.

\begin{figure}[t]
    \begin{minipage}{\linewidth}
        \centering\centering
        \includegraphics[width=\linewidth]{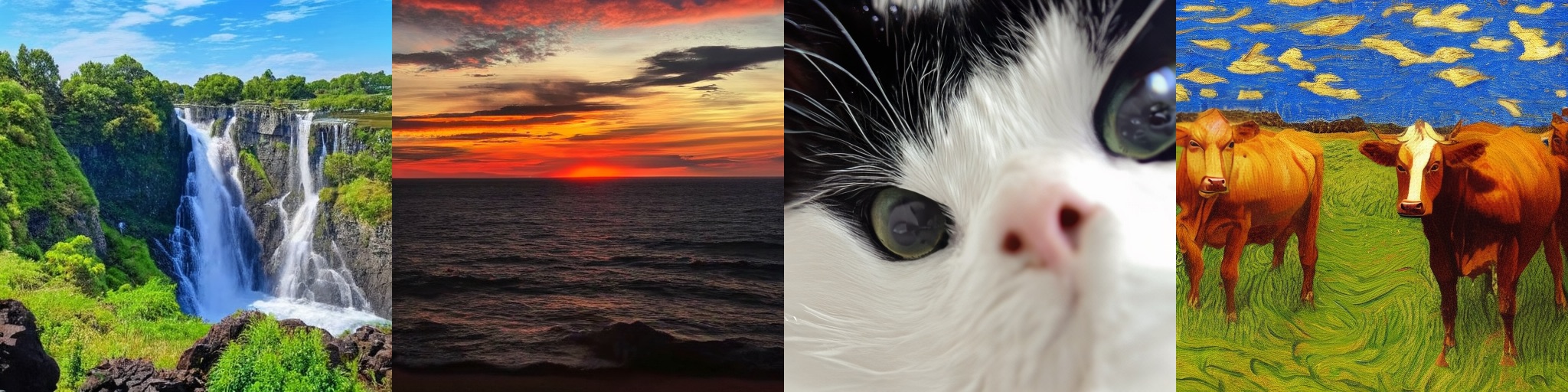}
        \includegraphics[width=\linewidth]{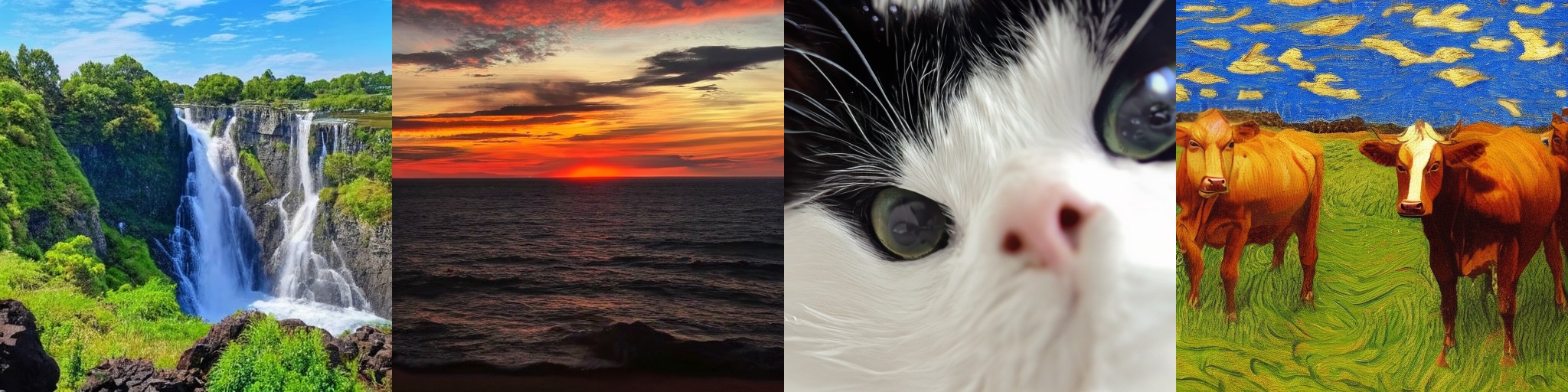}
        \includegraphics[width=\linewidth]{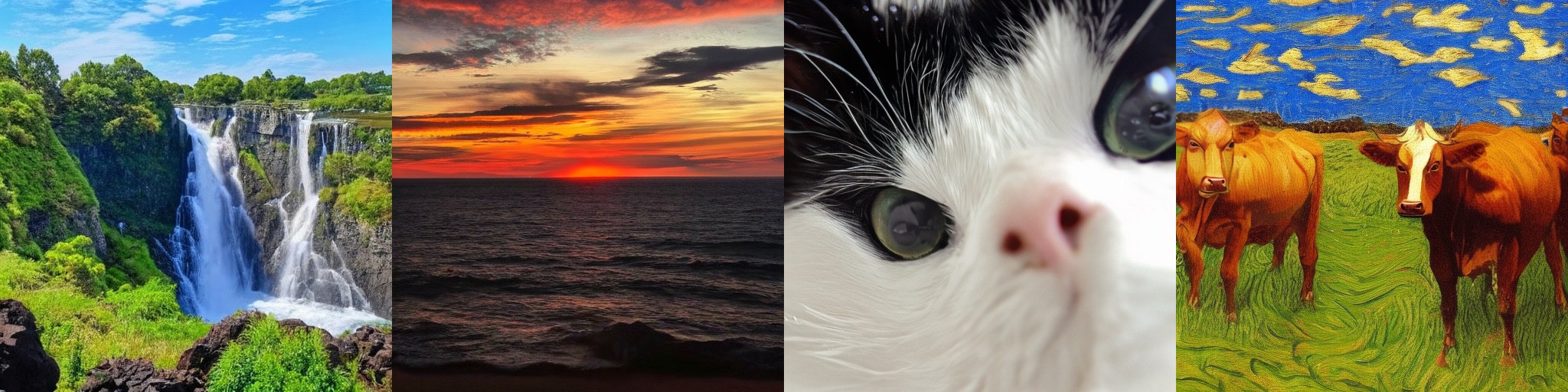}
        \includegraphics[width=\linewidth]{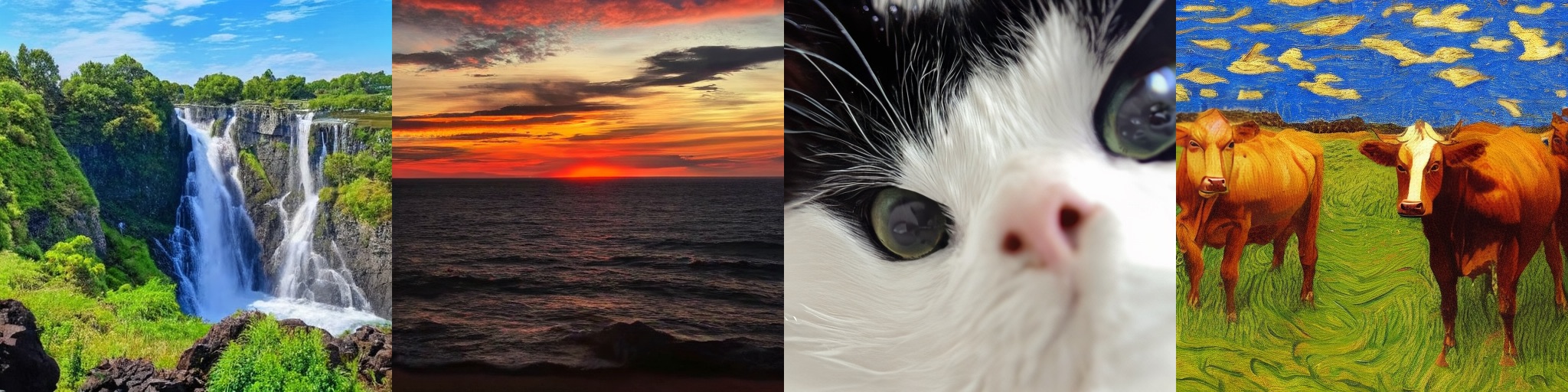}
        \includegraphics[width=\linewidth]{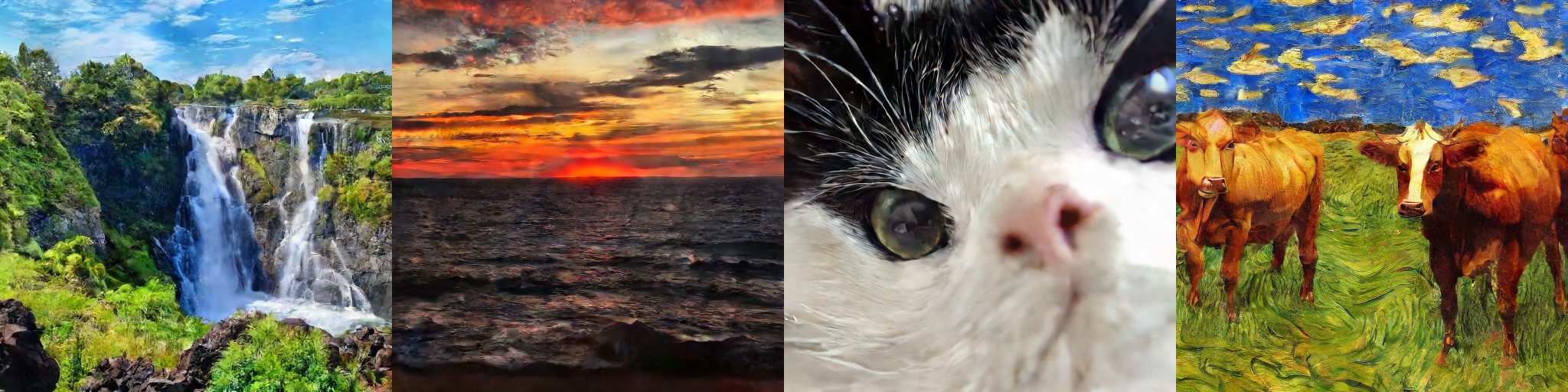}
        \includegraphics[width=\linewidth]{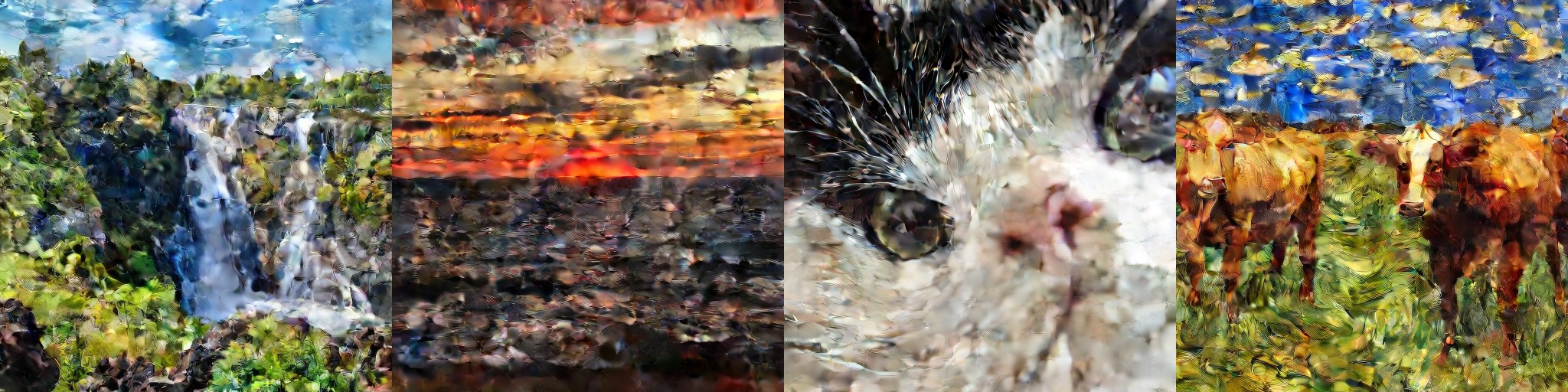}
    \end{minipage}
    \vspace{-0.5em}
    \caption{
    \textbf{Noisy Latents:}
    For a fixed set of latents corresponding to images generated with an LDM, we apply minute amounts of noise. From top to bottom, we illustrate the original image followed by addition of Gaussian noise with means of 0.01, 0.05, 0.1, 0.5, and 1.0 respectively (unit variance) to the orignal latents (first row). Note how at low noise levels (i.e. small variations in latents) there is no visible differences between generated images. The images generated correspond to text prompts for 
    \texttt{"View of a waterfall."},
    \texttt{"A beautiful view of the sunset against the ocean."},
    \texttt{"A close up shot of a black and white cat."}, and
    \texttt{"A Van Gogh style picture of cattle in a farm."}
    for columns left to right respectively.
    }
    \label{fig:app_noisy_latents}
\end{figure}

\section{Latent Adversarial Loss Detail}
\label{app:ladv}
Our adversarial loss in latent space is motivated by the nature of the latents we model: slight perturbations to the latents do not cause significant changes in the corresponding pixel-space image (illustrated in \Cref{fig:app_noisy_latents}). At the same time, operating in latent space is more efficient during training (compared to pixel space). A similar latent space adversarial loss is explored in \cite{Lin2024SDXLLightningPA}, however they rely on a large discriminator in the form of a pre-trained U-Net model and require several hand-crafted designs to enable stable training. In contrast, we use a simple discriminator network containing three convolutional layers.

In detail, our three convolutional layers are applied over the latents in the form of a set of prediction heads for each spatial location (motivated by \cite{Sauer2023StyleGANTUT,Sauer2023AdversarialDD}. The three convolutional layers contain 64, 64, 128 channels respectively, are implemented as spectral convolutions following \cite{Sauer2023StyleGANTUT} followed by batchnorm operations. A final point-wise convolution is applied to project the channel dimension to 1 (this output still contains spatial dimensions). We calculate this for both generated and natural images, and each of the outputs (corresponding to spatial locations) of our discriminator are mapped to 0 or 1 following our adversarial loss terms in \Cref{eq:natural_image_loss}.

We note that our adversarial loss is applied on fully generated images only (i.e. the final output of \modelname or natural images), in contrast to approaches such as \cite{Sauer2023AdversarialDD,Lin2024SDXLLightningPA} where intermediate stage latents (corresponding to partially corrupted images) are also used as targets with the adversarial loss. 

\begin{table}[t]
\small
\centering
\caption{
\textbf{Additional Diversity Evaluations:} 
We report Vendi Score and FID score along with their percentage drop relative to teacher model ($^\dagger$ indicates teacher models). 
Compared against state-of-the-art temporal distillation methods ADD \cite{Sauer2023AdversarialDD} and DMD-v2 \cite{Yin2024ImprovedDM}, our model displays no loss in diversity. 
On the other hand, SDXL-Lightening \cite{Lin2024SDXLLightningPA} similarly demonstrates no diversity loss. However, our \modelname retains image quality (FID) much better while also retaining a similar level of diversity. 
Vendi scores are calculated on the 1632 Parti Prompts as described in \Cref{app:diversity}. In this evaluation only, we calculate FID scores on a \underline{10K subset of images} from the COCO 2014 dataset following \cite{Yin2024ImprovedDM,Lin2024SDXLLightningPA} to allow direct comparison to their reported results. Note that the reported FID numbers are borrowed directly from prior work (as reported in their work) except for LDM and ours. 
}\label{app:tbl_diversity}
\vspace{-0.5em}
\def\arraystretch{1.2}  
\setlength\tabcolsep{1.0em}  
\scalebox{0.85}{
\begin{tabular}{l|c|l|l}
\toprule
Method  & Steps & \text{Vendi Score $\uparrow$} & FID $\downarrow$ \\ \midrule
SDXL \cite{Podell2023SDXLIL} $^\dagger$        &  50     & 3.01             & 19.36 \\ 
ADD  \cite{Sauer2023AdversarialDD}             &   4     & 1.69\dec{43.9\%} & 23.19\dec{19.8\%} \\ 
DMD-v2 \cite{Yin2024ImprovedDM}                &   4     & 2.73\dec{9.34\%} & 19.32\inc{0.21\%} \\ 
SDXL-Lightening  \cite{Lin2024SDXLLightningPA} &   4     & 3.06\inc{1.66\%} & 24.46\dec{26.3\%}  \\ \midrule
LDM-L \cite{Rombach2021HighResolutionIS} $^\dagger$ & 50 & 2.82 & 16.52                  \\ \rowcolor{Gray}
\modelname-L (ours)                            &  33     & 2.82\inc{0.00\%}   & 15.60\inc{5.57\%} \\ 
\bottomrule
\end{tabular}}
\end{table}

\begin{figure*}[t]
    \begin{minipage}{\linewidth}
        \centering\centering
        \includegraphics[width=\linewidth]{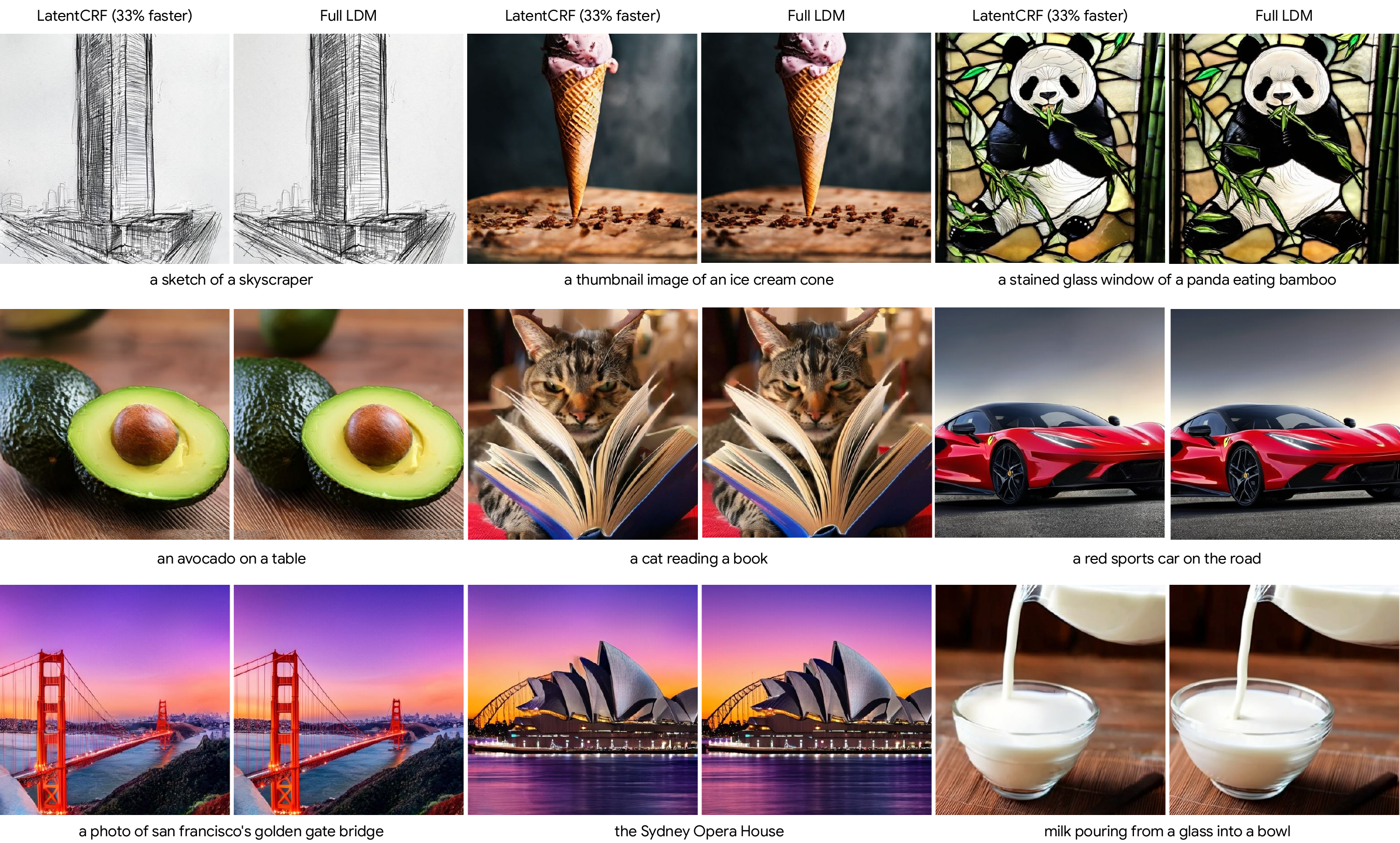}
    \end{minipage}
    \vspace{-0.5em}
    \caption{
    \textbf{Qualitative Results:} 
    Within each set of two, LatentCRF (left) speeds up LDM (right) by 33\% while maintaining image quality.
    In the top two rows we illustrate examples where our \modelname generated images exhibit visual quality competitive with the LDM baseline. In the bottom row, we illustrate a few failure cases of \modelname in comparison to LDM (slightly lower visual quality apparent on careful examination).
    }
    \label{app:qualitative_res1}
\end{figure*}

\section{Diversity Evaluation}
\label{app:diversity}
Diversity is an important facet of image generation and a key strength of diffusion models \cite{Xiao2021TacklingTG}. In boosting inference speeds of text-to-image diffusion, it is crucial to pay careful attention to the diversity of generated images. By diversity, we refer to the ability of models to generate distinct images for distinct starting noise vectors. This aspect of models remains relatively under explored, especially in quantitative terms. 

Motivated by this, we first construct a formal metric to evaluate the diversity of a given text-to-image model. We select the Vendi score \cite{Friedman2022TheVS} which is an established metric for evaluating diversity of data distributions, including in the visual domain \cite{Pasarkar2023CousinsOT}. In fact, prior work explores how such a metric can be adopted to measuring diffusion based image generation diversity using strong feature extractors \cite{Burgert2023DiffusionIH}. However, for text-to-image domain, to the best of our knowledge, there is no established benchmark for diversity evaluation. 

Motivated by this, we construct a benchmark for evaluating the diversity of a given text-to-image model. We utilize the Parti Prompts \cite{yu2022scaling} containing 1632 distinct image descriptions constructed specifically for text-to-image generation evaluation. For each caption (image description) we generate 16 images using 16 distinct noise vectors. We then use a CLIP vision encoder (ViT-L/14 336px) \cite{radford2021learning} to extract visual features for each of the 16 images followed by calculating their pairwise similarity to obtain a $16 \times 16$ similarity matrix. This similarity matrix is used to calculate a Vendi score following the implementation in \cite{Friedman2022TheVS}. This calculated score corresponds to a single sample (image description) and we average these values over all 1632 samples to obtain the final Vendi score diversity metric. This metric is reported in \Cref{table:results_main,table:diversity} in the main paper. 

We perform additional evaluations on two concurrent works \cite{Yin2024ImprovedDM,Lin2024SDXLLightningPA} and report these results in \Cref{app:tbl_diversity}. In comparison to DMD-v2 \cite{Yin2024ImprovedDM}, our approach retains diversity of the teacher model much better. SDXL-Lightning \cite{Lin2024SDXLLightningPA} exhibits similar diversity as our \modelname, but we outperform them on retaining image quality.

\begin{figure*}[t]
    \begin{minipage}{\linewidth}
        \centering\centering
        \includegraphics[width=\linewidth]{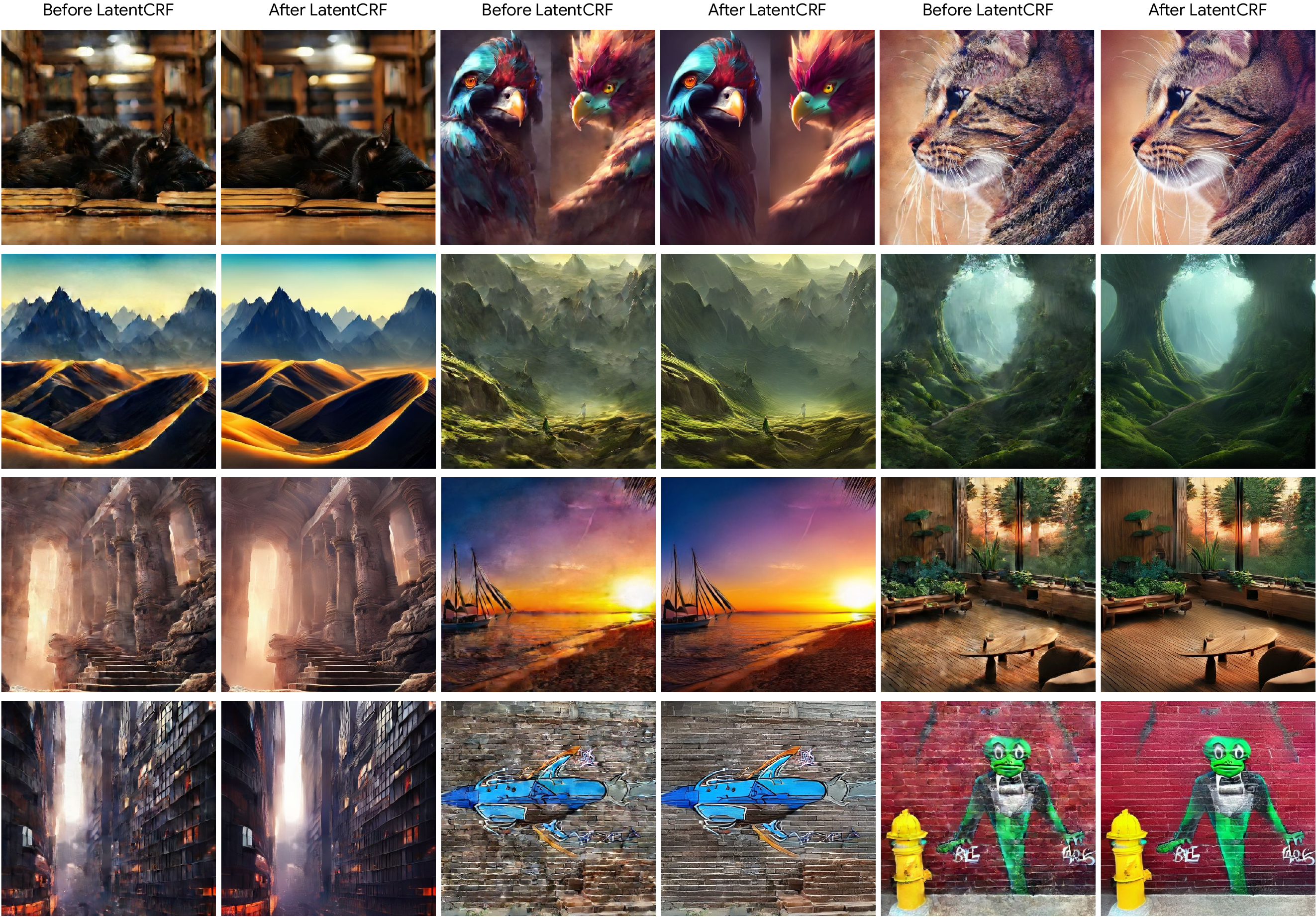}
    \end{minipage}
    \vspace{-0.5em}
    \caption{
    The figure shows the effect of applying our proposed lightweight Conditional Random Field (CRF) in the latent space of a Latent Diffusion Model (LDM). By replacing several LDM iterations with a single LatentCRF inference, we are able to increase inference speed by 33\%, while maintaining image quality and generation diversity. Each pair contains images that are the input (left) and output (right) of our \modelname. 
    }
    \label{app:qualitative_res2}
\end{figure*}

\section{More Qualitative Evaluations}
\label{app:qualitative}
We present more qualitative evaluations for our \modelname in \Cref{app:qualitative_res1,app:qualitative_res2}. We first illustrate image pairs generated from our \modelname and the LDM baseline, including failure cases of our model. These are presented in \Cref{app:qualitative_res1}. We also visualize the effect of applying our \modelname, presenting pairs of images containing input and output of our model. These are presented in \Cref{app:qualitative_res2}. 

\begin{figure*}[t]
    \begin{minipage}{0.03\linewidth}
    \centering
    \begin{tabular}{c}
         \rotatebox{90}{\texttt{SDXL}$^\dagger$}  \\[3.2em] 
         \rotatebox{90}{\texttt{ADD}}             \\[3.2em] 
         \rotatebox{90}{\texttt{SDXL-L}}          \\[3.2em] 
         \rotatebox{90}{\texttt{DMD}}             \\[3.2em] 
         \rotatebox{90}{\texttt{LDM}$^\dagger$}   \\[3.2em] 
         \rotatebox{90}{\texttt{Ours}}  \\
    \end{tabular}
    \end{minipage}
    \begin{minipage}{0.96\linewidth}
        \centering\centering
        \includegraphics[width=\linewidth]{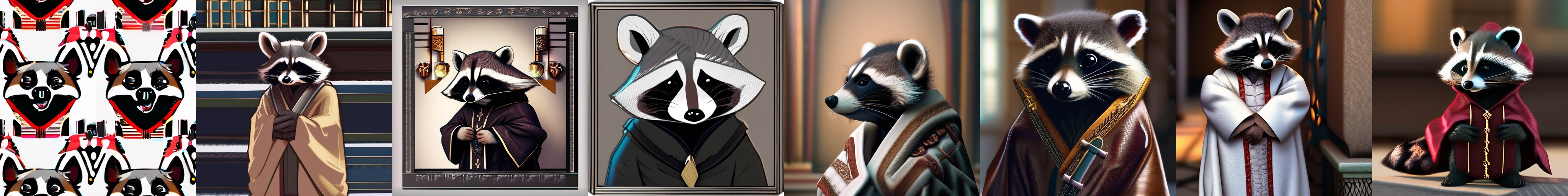}
        \includegraphics[width=\linewidth]{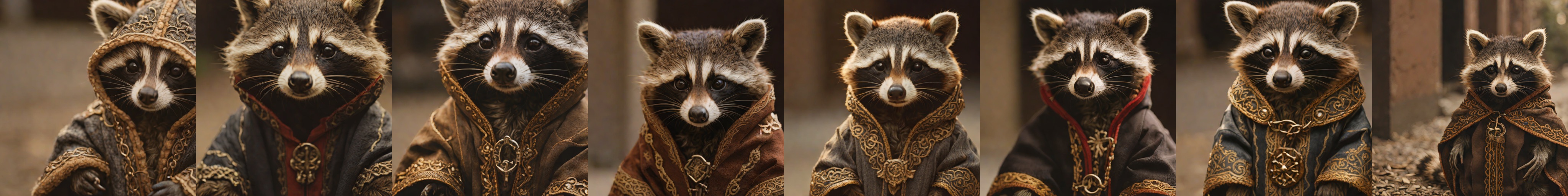}
        \includegraphics[width=\linewidth]{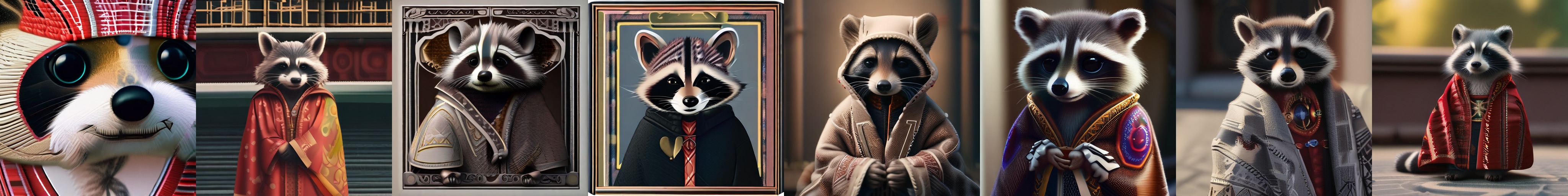}
        \includegraphics[width=\linewidth]{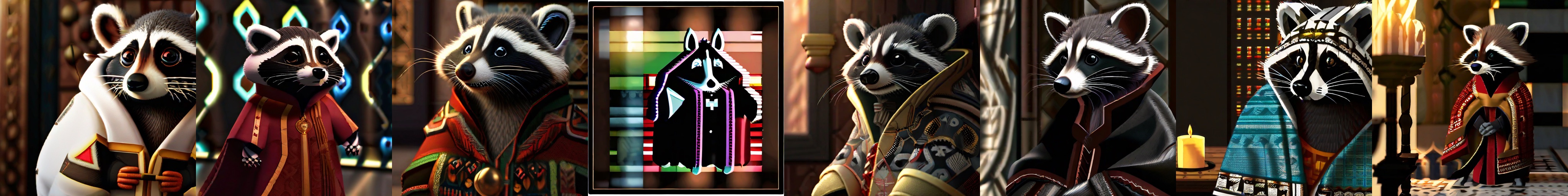}
        \includegraphics[width=\linewidth]{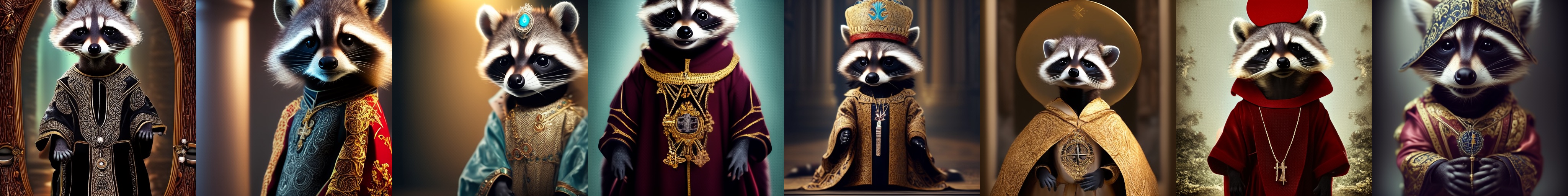}
        \includegraphics[width=\linewidth]{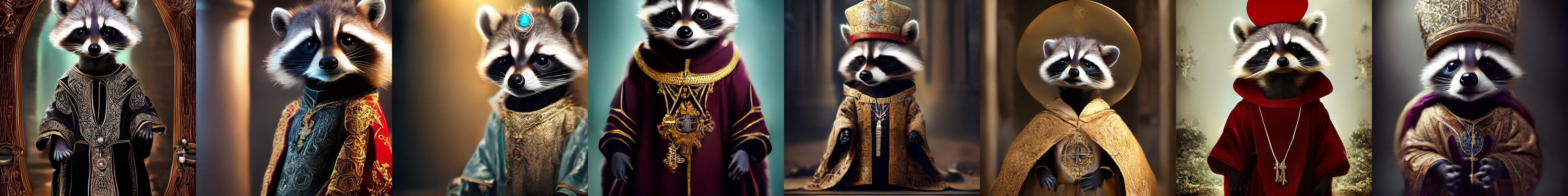}
    \end{minipage}
    \vspace{-0.5em}
    \caption{
    \textbf{Additional Diversity Visualization:}
    We visualize images generated for the prompt \texttt{`A cinematic shot of a baby racoon wearing an intricate italian priest robe.'} using a fixed set of noise vectors (common for all models). Each method uses some distillation from a teacher model (indicated with $^\dagger$). We compare against ADD \cite{Sauer2023AdversarialDD}, SDXL-Lightening \cite{Lin2024SDXLLightningPA} (SDXL-L), and DMD-v2 \cite{Yin2024ImprovedDM} (DMD). All of these models are distilled from SDXL (first row). 
    Our method is distilled from LDM, which is an internal implementation of \cite{Rombach2021HighResolutionIS} that is comparable to SDXL. 
    We highlight how our method retains both the diversity and visual quality of images in comparison to the teacher model. In fact, our \modelname generates almost the same images as the teacher model given common starting noise, while generating these images 33\% faster. 
    }
    \label{fig:app_diversity1}
\end{figure*}

\begin{figure*}[t]
    \begin{minipage}{0.03\linewidth}
    \centering
    \begin{tabular}{c}
         \rotatebox{90}{\texttt{SDXL}$^\dagger$}  \\[3.2em] 
         \rotatebox{90}{\texttt{ADD}}   \\[3.2em] 
         \rotatebox{90}{\texttt{SDXL-L}}\\[3.2em] 
         \rotatebox{90}{\texttt{DMD}}   \\[3.2em] 
         \rotatebox{90}{\texttt{LDM}$^\dagger$}   \\[3.2em] 
         \rotatebox{90}{\texttt{Ours}}  \\
    \end{tabular}
    \end{minipage}
    \begin{minipage}{0.96\linewidth}
        \centering\centering
        \includegraphics[width=\linewidth]{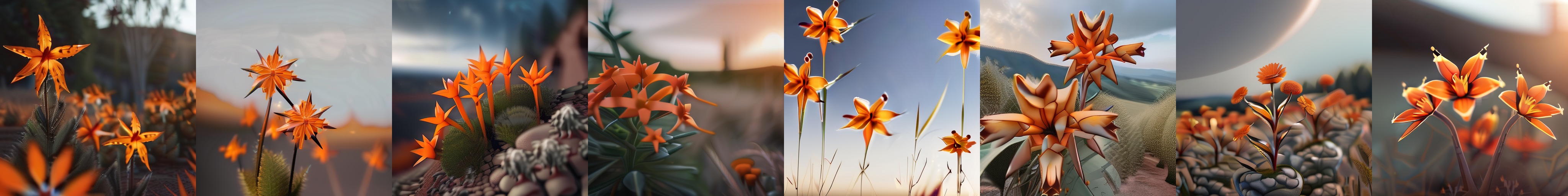}
        \includegraphics[width=\linewidth]{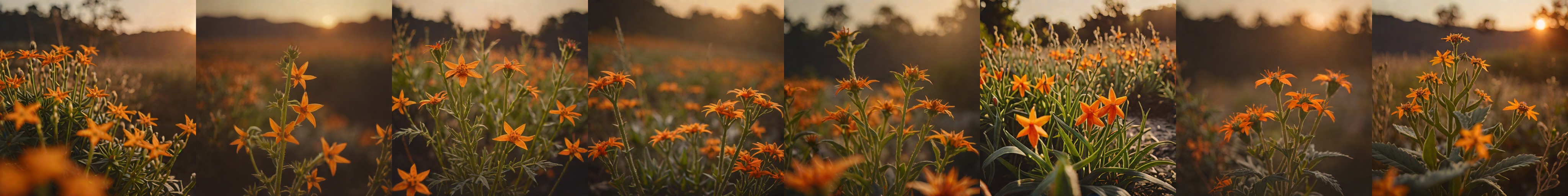}
        \includegraphics[width=\linewidth]{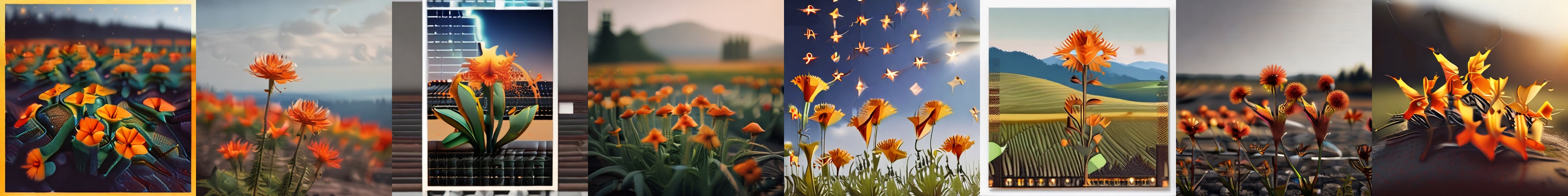}
        \includegraphics[width=\linewidth]{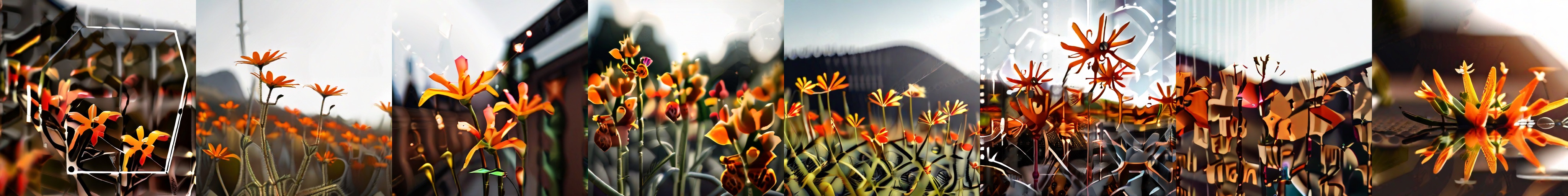}
        \includegraphics[width=\linewidth]{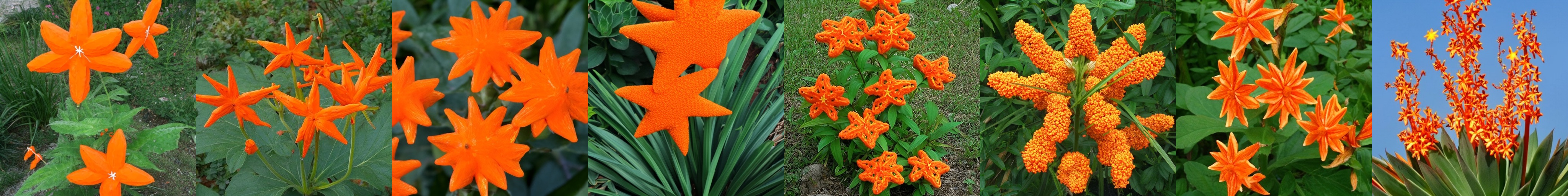}
        \includegraphics[width=\linewidth]{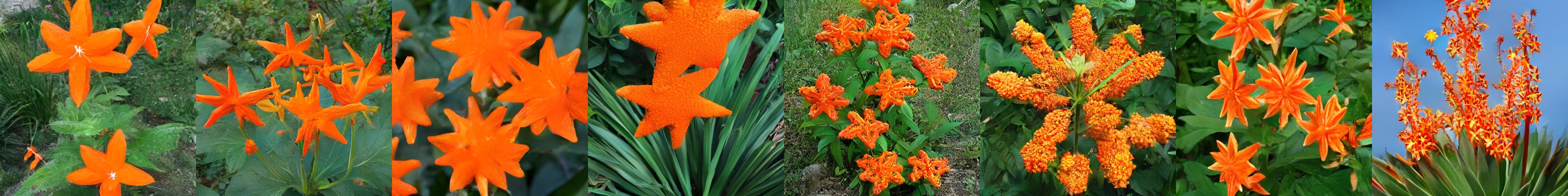}
    \end{minipage}
    \vspace{-0.5em}
    \caption{
    \textbf{Additional Diversity Visualization 2:}
    We visualize more images following the same settings in \Cref{fig:app_diversity1}. We use the caption \texttt{`A plant with orange flowers shaped like stars.'} to generate images. While our LDM demonstrates strong visual quality and diversity, our \modelname retains both of these much better than alternate methods with the benefit of being 33\% faster during inference. 
    }
    \label{fig:app_diversity2}
\end{figure*}

We present two additional visualizations for diversity of generated images in \Cref{fig:app_diversity1,fig:app_diversity2}. We include some failure cases of our model as well (i.e. where generated image quality is visually subpar to LDM). However, these visualizations indicate that on average our \modelname outperforms prior distillation works in terms of image quality and diversity retained from the teacher model. This is consistent with the quantitative evaluations presented in \Cref{app:tbl_diversity}.

\section{LDM Schedule for \modelname}
\label{app:sparse_schedules}
Our default setting for \modelname utilizes an LDM schedule with 40 DDIM timesteps. 
For a fixed LDM, using sparser reverse diffusion schedules leads to faster generation but at reduced image quality. In our case, as validated in \Cref{sec:exp}, our \modelname is able to modify intermediate latents from a 40 timestep schedule and generate images of quality matching a 50 timestep schedule. 

\begin{table}[t]
\small
\centering
\caption{
\textbf{LDM Schedule in \modelname Pipeline:} 
We explore integrating our \modelname with sparser LDM schedules for further speed-up and uncover promising results.
Using a schedule containing only 20 LDM iterations (calls to U-Net), we still achieve competitive image quality at over 50\% faster inference speed. 
}
\label{table:app_ablate_schedule}
\vspace{-1.0em}
\def\arraystretch{1.2}  
\setlength\tabcolsep{1.2em}  
\scalebox{0.88}{
\begin{tabular}{l|c|c|l}
\toprule
Method   & Iter & FID $\downarrow$ & \text{Speed (ms)} $\downarrow$ \\ \midrule
LDM \cite{Rombach2021HighResolutionIS}  &  50  & 12.75  & 782.1  \\
LDM \cite{Rombach2021HighResolutionIS}  &  33  & 12.78  & 501.8\inc{35.9\%}  \\
LDM \cite{Rombach2021HighResolutionIS}  &  20  & 12.80  & 323.9\inc{58.6\%}  \\ \midrule 
\modelname (ours)                       &  33  & 11.58  & 523.4\inc{33.2\%}  \\ 
\modelname (ours)                       &  20  & 11.67  & 344.7\inc{55.9\%}  \\ 
\bottomrule
\end{tabular}}
\end{table}



We now investigate if even more sparse schedules could be used with our \modelname and report results in \Cref{table:app_ablate_schedule}. We build a variant that uses only 20 LDM iterations, where we first run 19 LDM steps, apply our CRF, and conclude with a final LDM step. This varaint achieves an impressive 50\% inference speedup. While the quantitative results are on par with our full \modelname and with the LDM, upon close examination of qualitative results we do occasionally see more significant artifacts.


\section{Forward Diffusion Process}
\label{app:forward_diffusion}
Diffusion models are trained by generating targets for intermediate timesteps using a forward diffusion process. In LDM \cite{Rombach2021HighResolutionIS}, forward diffusion is applied within latent space. Given a natural image and its latent $\vz_0$, a corrupted latent $\vz_t$ for a timestep $t$ is generated by applying Gaussian noise to the original latent $\vz_0$. In our CRF formulation, considering how the observed noisy input and true underlying value correspond to $\vz_t$ and $\vz_0$. Therein, the difference between observed and actual data, $\vz_0 - \vz_t$ corresponds to a combination of Gaussian noise. We refer the reader to \cite{Ho2020} for further details on the forward diffusion process.

\section{Number of Iterations in CRF Inference}
\label{app:num_iterations}
Our CRF inference algorithm contains a \texttt{num\_iterations} parameter that controls its convergence. We discuss this parameter and evaluate its convergence in \Cref{app:mrf_converge} along with visualizations in \Cref{fig:app_ldm_converge}. In our \modelname, we fix this parameter to 5 during both training as well as inference. 





\end{document}